\documentclass[11pt, oneside]{article}
\usepackage{geometry}
\usepackage{graphicx}				
\usepackage{setspace}							
\usepackage{amssymb}
\usepackage{amsmath}
\usepackage{amsfonts}
\usepackage{float}
\usepackage{theoremref}
\usepackage{hyperref}
\usepackage{appendix}
\usepackage[colorinlistoftodos]{todonotes}
\usepackage{cleveref}
\usepackage[utf8]{inputenc}
\usepackage{mathtools}
\usepackage{abstract}
\usepackage{tikz}
\usepackage{algorithm}
\usepackage{algorithmic}
 
\usepackage{amsthm}

\newtheorem{theorem}{Theorem}

\theoremstyle{plain}
\newtheorem{proposition}{Proposition}

\newtheorem{lemma}{Lemma}
\newtheorem{corollary}{Corollary}

\theoremstyle{plain}

\theoremstyle{plain}
\newtheorem{definition}{Definition}

\theoremstyle{remark}

\theoremstyle{definition}

\usetikzlibrary{bayesnet}
\usetikzlibrary{fit}
\DeclareMathOperator*{\argmin}{arg\,min}

\DeclareMathOperator{\erfc}{erfc}
\DeclareMathOperator{\erf}{erf}

\DeclareMathOperator*{\prox}{Prox}
\DeclarePairedDelimiter\abs{\lvert}{\rvert}%
\DeclarePairedDelimiter\norm{\lVert}{\rVert}%
\newcommand*\samethanks[1][\value{footnote}]{\footnotemark[#1]}

\begin{document}
\title{Asymptotic errors for convex penalized linear regression beyond Gaussian matrices}
\date{}
\author{Cédric Gerbelot\thanks{Laboratoire de Physique de l’Ecole normale sup\'erieure, ENS, Universit\'e PSL, CNRS, Sorbonne Université, Universit\'e de Paris, F-75005 Paris, France}, Alia Abbara\samethanks \thickspace \thickspace and Florent Krzakala\samethanks}
\maketitle 

\begin{abstract}%
    We consider the problem of learning a coefficient vector $\bf x_0 \in \mathbb R^N$ from noisy linear observations $\mathbf{y} = \mathbf{F}{\mathbf{x}_{0}}+\mathbf{w} \in \mathbb R^M$ in the high dimensional limit $M,N \to \infty$ with $\alpha \equiv M/N$ fixed. We provide a rigorous derivation of an explicit formula ---first conjectured using heuristic methods from statistical physics--- for the asymptotic mean squared error obtained by penalized convex regression estimators such as the LASSO or the elastic net, for a class of very generic random matrices corresponding to rotationally invariant data matrices with arbitrary spectrum. The proof is based on a convergence analysis of an oracle version of vector approximate message-passing (oracle-VAMP) and on the properties of its state evolution equations. Our method leverages on and highlights the link between vector approximate message-passing, Douglas-Rachford splitting and proximal descent algorithms, extending previous results obtained with i.i.d. matrices for a large class of problems. We illustrate our results on some concrete examples and show that even though they are asymptotic, our predictions agree remarkably well with numerics even for very moderate sizes.
\end{abstract}
\section{Introduction}

Solving a regression problem with convex penalty in high dimension is certainly one of the most fundamental questions in a number of disciplines, ranging from statistical learning to signal processing. We shall consider this standard quadratic minimization problem on a given input space $\mathcal{X} \subset \mathbb{R}^{N}$ with $M$ samples:
\begin{equation}
    \mathbf{x}^{*} = \argmin_{\mathbf{x}\in\mathbb{R}^{N}} \frac{1}{2} \norm{\mathbf{y}-\mathbf{F}\mathbf{x}}_{2}^{2}+f(\mathbf{x})
    \label{main-problem}
\end{equation}
where $\mathbf{F} \in \mathbb{R}^{M \times N}$ is a known data matrix (in statistics/machine learning \cite{graybill1976theory}) or a a known measurement matrix (in signal processing/compressed sensing \cite{donoho2006compressed}), and $f$ a proper, closed, convex and separable regularization function. Examples includes ridge Regression \cite{marquardt1975ridge}, the LASSO \cite{tibshirani1996regression}, or Elastic nets \cite{zou2005regularization}. Here, we assume the  vector $\mathbf{y}$ has been obtained according to a noisy linear process as
\begin{equation}
    \mathbf{y} = \mathbf{F}{\mathbf{x}_{0}}+\mathbf{w}
    \label{main-y}
\end{equation}
where all elements from the vector $\mathbf{x}_{0} \in \mathbb{R}^{N}$ are identically and independently distributed (i.i.d.) according to an arbitrary given distribution $\phi_{0}(x_{0})$, and $\mathbf{w} \in \mathbb{R}^{M}$ is an i.i.d.  Gaussian white noise of zero mean and variance $\Delta_{0}$, independent of $\mathbf{F}$ and $\mathbf{x}_{0}$. We shall present asymptotically exact expressions for the mean squared error on the recovery of $  \mathbf{x}_0$ , that is on the error
\begin{equation}
{\rm MSE} =  \mathbb{E}\left[ \norm{\mathbf{x}_{0}-\mathbf{x}^{*}}_{2}^{2}\right].
\label{def-errors}
\end{equation}
Our asymptotic result will hold almost surely for a specific type of random matrices: we shall consider sequences of random matrices $\mathbf{F}$ with fixed aspect ratio $\alpha \equiv M/N$ as $M,N\!\to\!\infty$. 

In a pioneering paper, \cite{bayati2011lasso} considered this case for matrices  $\mathbf{F}$ with independent Gaussian entries, and provided a rigorous derivation of an explicit formula for the asymptotic mean squared error of the LASSO estimator. Our goal here is to go beyond independent and Gaussian matrices, and to give instead an asymptotic formula for generic matrices. Our matrices $\mathbf{F}$ will be assumed to be rotationally invariant: their singular value decomposition can be written $\mathbf{F} = \mathbf{U}\mathbf{D}\mathbf{V}^{T}$ where $\mathbf{U}, \mathbf{V}$ are Haar distributed (i.e. uniformly sampled over the orthogonal group) and $\mathbf{D}$ is an arbitrary diagonal matrix containing the singular values of $\mathbf{F}$.  While this setting is certainly specific, there is a long standing tradition for such problems in signal processing \cite{rangan2009asymptotic}, statistical physics \cite{guo2005randomly}, random matrix theory \cite{guionnet2009asymptotics} and communications theory  \cite{tulino2004random}. From the point of view of statistical learning, this model allows to give "typical-case" results, that represent an alternative and appealing approach to the worst-case analysis \cite{mohri2018foundations}. Although the latter is both mathematically rigorous and robust, it sometimes fails to predict interesting behaviors, like the recently observed double descent effects in generalization performance curves in overparametrized neural networks \cite{belkin2019reconciling}. Over the last few decades a considerable body of theoretical work has been made on
"typical-case" scenarios, especially within the framework of statistical
mechanics \cite{seung1992statistical,watkin1993statistical,mezard2009information,advani2013statistical,zdeborova2016statistical} and this direction is currently witnessing a burst of activity, see e.g. \cite{sur2019likelihood,hastie2019surprises,mei2019generalization}.  

\subsection{Main contributions}
Our main contributions are the following:
\\
\begin{itemize}
    \item We provide an analytical formula for the reconstruction error of problem \eqref{main-problem} in the asymptotic setup, for all convex penalties (including for instance LASSO and Elastic net), for all rotationally invariant sequences of matrices $\mathbf{F}$, extending the results of 
    \cite{bayati2011lasso,el2013robust} beyond Gaussian matrices.
      \item In doing so, we give a mathematically rigorous proof of the replica formula obtained heuristically from statistical physics for this problem \cite{rangan2009asymptotic,kabashima2012typical,kabashima2014signal}. To the best of our knowledge, this is the first proof of such formulas for generic rotationally invariant matrices.
    \item Our proof analysis has an interest of its own, and builds on a detailed mapping between proximal descent algorithms and maximum a posteriori forms of message-passing algorithms. In particular, it gives
upper bounds on the convergence rates of an oracle version of vector approximate message-passing~\cite{rangan2019vector}, and discusses a simple method to enforce convergence.
    \item Finally, we also show a rigorous proof for the statistical distribution of the estimator defined by \eqref{main-problem} for sufficiently strongly convex $f$, and conjecture its validity for any convex $f$.
\end{itemize}

\subsection{Related works}

\paragraph{Asymptotic distribution of M-estimators and reconstruction error ---}
The initial effort to prove asymptotic reconstruction error in this type of problem originates in \cite{bayati2011lasso} for the LASSO with Gaussian matrices. Their proof is based on a sequence of specifically designed iterates, whose statistical properties are analytically tracked in an asymptotic setting. The key idea is built on a modification of the celebrated iterative soft thresholding algorithm (ISTA) \cite{daubechies2004iterative} inspired by statistical physics. This additional term allows the exact computation of the conditional expectation of key quantities at each iteration based on the $\sigma$-algebra generated by the previous observations. This idea, initially presented for spin glasses in \cite{bolthausen2009high}, was later transposed to statistics as a leave-one-out method for ridge regularized M-estimators in \cite{el2013robust}, among others. In a recent paper \cite{sur2019likelihood}, the probabilistic setting with a converging sequence is once again used to prove the statistical distribution of logistic regression problems. This technique, now part of the modern theoretical tools of learning theory and high-dimensional statistics, is yet to be extended to problems involving correlated matrices.

\paragraph{Replica heuristic and their rigorous proofs ---}
As the reconstruction error of problems akin to \eqref{main-problem} is of interest in the machine learning and signal processing literature, so are overlap parameters between estimators and underlying truth in statistical physics. In the statistical physics literature, these quantities are calculated using the reknown replica method \cite{mezard1987spin}, a powerful heuristic calculation of the log partition function of a Bayesian problem based on the identity $\log Z = \lim_{n\to 0} \frac{Z^{n}-1}{n}$. This method has led to a number of predictions in various fields of computer science \cite{mezard2009information}, and in particular in machine learning \cite{seung1992statistical,watkin1993statistical} regarding the generalization error of single neurons \cite{opper1996statistical}, support vector machines \cite{dietrich1999statistical} or fundamental quantities such as capacity of neural nets \cite{gardner1988space}. Although it is adaptable to many machine learning problems, see for example \cite{zdeborova2016statistical} for a review, the replica method suffers from the fact that it is a non-rigorous approach. While replica calculations are often restricted to i.i.d. problems, they have been extended to rotationally invariant matrices, see for instance \cite{kabashima2014signal}, using the Harish-Chandra-Itzykson-Zuber \cite{collins2003moments} formula. A substantial effort has been dedicated to prove the replica formula in specific settings, for problems originating in both statistical physics \cite{talagrand2003spin} and machine learning, such as low rank matrix factorization \cite{dia2016mutual}. Generic methods have also been proposed based on the Guerra interpolation technique \cite{guerra2003broken}, and later extended to modern Bayesian inference \cite{barbier2019adaptive}. In particular, rigorous proof for the replica formula in Bayes MMSE version of solving the inverse problem (\ref{main-y}) has been given in \cite{barbier2016mutual,reeves2016replica,barbier2019optimal} for Gaussian matrices, and in \cite{barbier2018mutual} for (a large part of) rotationally invariant matrices. 

\paragraph{Message-passing algorithms ---}
Both the replica approach and ongoing body of work on the asymptotic distribution of M-estimators can be linked to variational inference \cite{wainwright2008graphical} through approximate message-passing algorithms \cite{mezard2009information}. This family of algorithms is a statistical physics inspired variant of belief propagation, where local beliefs are approximated by Gaussian distributions. In a probabilistic framework, this leads to a powerful alternative to Monte Carlo methods that scale well with data dimension. One of the key properties of these algorithms are the so-called state evolution equations, an iterative scalar equivalent model which allows to track the asymptotic statistical properties of the iterates. A series of groundbreaking papers initiated with \cite{bayati2011dynamics} proved the exactness of these equations in the large system limit, and extended the method to treat nonlinear problems \cite{rangan2011generalized} and handle rotationally invariant matrices \cite{rangan2019vector}. The latter involves vector approximate message-passing, and lies at the center of our approach.
\par
The maximum a posteriori (MAP) forms of approximate message-passing algorithms are closely linked to proximal descent methods \cite{parikh2014proximal}. As pointed out above, the original MAP approximate message-passing amounts to writing the ISTA with a second order correction term \cite{montanari2012graphical} based on mean values of previous iterates, sometimes referred to as the Onsager reaction term in theoretical physics. Vector-approximate message-passing and the related class of expectation consistent inference algorithms \cite{minka2001family}, \cite{opper2005expectation}; \cite{fletcher2016expectation} yield adaptative versions of Douglas-Rachford/ADMM \cite{parikh2014proximal} descent methods where the step sizes match the local curvature of the cost function at each iteration. The overall result is a family of faster algorithms than proximal descent ones, 
with asymptotically exact analytical forms for the statistical properties of the iterates. Their main drawback is the restricted class of matrices to which they are applicable, 
and the somewhat more complicated structure they present due to the adaptative terms. Although this impedes message-passing methods from becoming mainstream optimization procedures, they remain formidable theoretical tools.

\section{Main Results}
 Our main result is a rigorous proof of a replica conjecture left open by \cite{rangan2009asymptotic,vehkapera2016analysis}. Here we state our main theorems, show how they agree with simulations at finite size and give a brief sketch of proof. The framework of assumptions is the same as the one introduced in \cite{bayati2011dynamics} and later used in \cite{rangan2019vector}. We provide details on this framework in appendix \ref{appendix:analysis_framework}.

\subsection{Main theorem}
\begin{theorem}
\thlabel{theorem1}
Consider problem \eqref{main-problem} with a proper closed, convex and separable $f$. Consider that the empirical distributions of the underlying truth $\mathbf{x}_{0}$ and singular values of the rotationally invariant sensing matrix respectively converge with second order moments, as defined in appendix \ref{appendix:analysis_framework}, to given distributions $p_{X_{0}}$ and $p_{S}$. Assume that the distribution $p_{S}$ is non-trivial and has compact support. Finally consider the limit $M,N \to \infty$ with fixed ratio $M/N = \alpha$. Then the average mean squared error ${\rm MSE} = \frac{1}{N} \mathbb{E}\left[ \norm{\mathbf{x}_{0}-\mathbf{x}^{*}}_{2}^{2}\right]$ for the estimator prescribed by \eqref{main-problem} is given by the fixed point $\tilde E$ of the equations:
\begin{footnotesize}
\begin{subequations}
\label{main-result}
\begin{align}
    \tilde{V} &= \mathbb{E}\left[\dfrac{1}{\mathcal{R}_{\mathbf{C}}(-\tilde{V})}\mbox{$\prox'_{f/ \mathcal{R}_{\mathbf{C}}(-\tilde{V})}\left(x_{0} + \dfrac{z}{\mathcal{R}_{\mathbf{C}} \left( -\tilde{V}\right)} \sqrt{\left( \tilde{E} - \Delta_0 \tilde{V} \right)\mathcal{R}_{\mathbf{C}}' \left( -\tilde{V}\right) + \Delta_0 \mathcal{R}_{\mathbf{C}} \left( -\tilde{V}\right)}\right)$}\right] \label{V-replica}\\
    \tilde{E} &= \mathbb{E}\left[\left\lbrace \mbox{$\prox_{f/ \mathcal{R}_{\mathbf{C}}(-\tilde{V})}\left(x_{0} + \dfrac{z}{\mathcal{R}_{\mathbf{C}} \left( -\tilde{V}\right)} \sqrt{\left( \tilde{E} - \Delta_0 \tilde{V} \right)\mathcal{R}_{\mathbf{C}}' \left( -\tilde{V}\right) + \Delta_0 \mathcal{R}_{\mathbf{C}} \left( -\tilde{V}\right)}\right)-x_{0}$}\right\rbrace^{2}\right], \label{E-replica}
\end{align}
\end{subequations}
\end{footnotesize}
where $\mathbf{C} = \mathbf{F}^{T}\mathbf{F}$, $\mathcal{R}_{\mathbf{C}}$ is the R-transform with respect to the spectral distribution of $\mathbf{F}^T \mathbf{F}$, which is defined in appendix ~\ref{appendix : replica_full}, and expectations are over $z \sim \mathcal{N}(0,1)$ and $x_0 \sim p_{X_{0}}$. $\mbox{Prox}$ is the proximal operator defined as: \\
\begin{equation}
    \forall \gamma \in \mathbb{R}^{+}, x, y \in \mathbb{R} \quad  \mbox{$\prox_{\gamma f}$}(y) \equiv \argmin_{x} \left\lbrace f(x)+\frac{1}{2\gamma}(x-y)^{2} \right\rbrace.
\end{equation}
\quad \\
Additionally, for any instance of problem \eqref{main-problem}, consider the regularized problem where $f$ is replaced by $h = f+\frac{\lambda_{2}}{2}\norm{.}_{2}^{2}$ and $\mathbf{x}^{*}_{\lambda_{2}}$ the corresponding solution. We then have the following result on the element wise distribution of this solution:
\begin{small}
\begin{align}
  & \exists \thickspace \lambda_{2}^{*} \thickspace \mbox{s.t.} \thickspace \forall \thickspace \lambda_{2}>\lambda_{2}^{*} : \notag \\
    x^{*}_{\lambda_{2}} \sim \thickspace
    &\mbox{$\prox_{h/ \mathcal{R}_{\mathbf{C}}(-\tilde{V})}$} \left(x_0 + 
     \dfrac{z}{\mathcal{R}_{\mathbf{C}} \left( -\tilde{V}\right)} \sqrt{\left( \tilde{E} - \Delta_0 \tilde{V} \right)\mathcal{R}_{\mathbf{C}}' \left( -\tilde{V}\right) + \Delta_0 \mathcal{R}_{\mathbf{C}} \left( -\tilde{V}\right)} \right).
     \label{distrib}
\end{align}
\end{small}
\end{theorem}

For completeness, the replica computation leading to \eqref{main-result}, that appeared in \cite{rangan2009asymptotic,vehkapera2016analysis}, is given in appendix \ref{appendix : replica_full} where we used the notations of \cite{krzakala2012probabilistic}. While we believe and conjecture that the second property \eqref{distrib} holds for any $\lambda_{2} \geqslant 0$, we could prove it for an open subset of $\lambda_2$. The asymptotic error from \thref{theorem1} can be equivalently written:
\begin{theorem}
\thlabel{theorem2}
Under the assumptions presented above, the average mean squared error of  $\mathbf{x}^{*}$ is equivalently given by the fixed point of the state evolution equations of vector approximate message-passing \cite{rangan2019vector}.
\begin{subequations}
\label{VAMP_se}
\begin{align}
\alpha_{1k} &= \mathbb{E}\left[\mbox{$\prox'_{\frac{1}{A_{1k}}f}(x_{0}+P_{1k})$}\right] \quad V_{1k} = \frac{\alpha_{1k}}{A_{1k}} \\
A_{2k} &= \frac{1}{V_{1k}}-A_{1k} \hspace{2.8cm} \tau_{2k} = \frac{1}{(1-\alpha_{1k})^{2}}\left[\mathcal{E}_{1}(A_{1k},\tau_{1k})-\alpha_{1k}^{2}\tau_{1k}\right] \\
\alpha_{2k} &= \mathbb{E}\left[\frac{A_{2k}}{\lambda_{\mathbf{C}}+A_{2k}}\right] \hspace{2.05cm} V_{2k} = \frac{\alpha_{2k}}{A_{2k}} \\
A_{1,k+1} &= \frac{1}{V_{2k}}-A_{2k} \hspace{2.35cm} \tau_{1,k+1} = \frac{1}{(1-\alpha_{2k})^{2}}\left[\mathcal{E}_{2}(A_{2k},\tau_{2k})-\alpha_{2k}^{2}\tau_{2k}\right]
\end{align}
\end{subequations}
where $\mathcal{E}_{1}$ and $\mathcal{E}_{2}$ are function defined  by:
\begin{small}
\begin{equation}
    \mathcal{E}_{1}(A_{1k},\tau_{1k}) = \mathbb{E}\left[\left( \mbox{$\prox$}_{\frac{1}{A_{1k}}f}(x_{0}+P_{1k})-x_{0} \right)^{2}\right]\, , \quad
   \mathcal{E}_{2}(A_{2k},\tau_{2k}) = \mathbb{E}\left[\frac{\Delta_{0}\lambda_{\mathbf{C}}+\tau_{2k}A_{2k}}{(\lambda_{\mathbf{C}}+A_{2k})^{2}}\right]\, \label{equation:e2_eig} 
\end{equation}
\end{small}
and
\begin{equation}
    P_{1k} \sim \mathcal{N}(0,\tau_{1k}).
\end{equation}
At the fixed point, $\mathcal{E}_{1} = \mathcal{E}_{2} = \frac{1}{N}\mathbb{E}\left[ \norm{\mathbf{x}_{0}-\mathbf{x}^{*}}_{2}^{2}\right]$.
\label{second-res}
\end{theorem}

When solving these equations numerically with an explicit distribution of singular values, we found that \thref{theorem2} was sometimes better suited for numerical evalutation than \thref{theorem1} as it gives a more stable numerical scheme upon (damped) iteration. Both theorems draw a strong connection between asymptotic formulas for the MSE of Problem \ref{main-problem} and the various transforms used in free probability and random matrix theory \cite{tulino2004random}.

\subsection{Applications and numerical experiments}
\label{subsection : experiments}

We compare the result of \thref{theorem2} with numerics on two typical problems. In both, the underlying truth vector $\bf x_{0}$ is an i.i.d. one where each element is pulled from a Gauss-Bernoulli distribution with sparsity parameter $\rho \in \mathbb{R_{+}}$:
\begin{equation}
    \phi_{0}(x_{0}) = (1-\rho)\delta(x_{0})+\rho\frac{1}{\sqrt{2\pi}}\exp{(-x_{0}^{2}/2)}\, ,
\end{equation}
and the training vector $\mathbf{y}$ is obtained from \eqref{main-y}. All experimental points are obtained using the Scikit-learn \cite{pedregosa2011scikit} implementation of the LASSO, which uses a coordinate descent method with duality gap convergence control \cite{friedman2010regularization,kim2007interior}.
\paragraph{Signal recovery with row orthogonal matrices ---} 
In the first model, we consider a setting popular in signal processing and use row orthogonal matrices. Such random matrices are very similar to subsampled Fourier and Hadamard matrices, and play a fundamental role in e.g. compressed sensing \cite{tsaig2006extensions} and communication \cite{guo2005randomly}. We aim to recover the underlying sparse vector using a LASSO regression and tune the regularization parameter. We want to compare the performance using an i.i.d. Gaussian matrix pulled from $\mathcal{N}(0,\frac{1}{N})$ and a row-orthogonal one, i.e. where the singular values of $\mathbf{F}$ are set to one, which gives the following distribution for the eigenvalues of $\mathbf{C}$:
\begin{equation}
    \lambda_{\mathbf{C}} \sim \max(0,1-\alpha)\delta(0)+\min(1,\alpha)\delta(1).
\end{equation}
We take $M,N = 200,100$ ($\alpha =2$), $\Delta_{0} = 0.01$ and $\rho = 0.3$. Each point is an average over $10^{4}$ realizations. The error bars in this case are vanishingly small ($\sim 10^{-5}$). We see that an excellent agreement is obtained with the asymptotics of \thref{theorem1} although the simulation matrices are rather small, indicating that the prediction remains very good at finite values of $M,N$.

\begin{figure}[H]
\begin{center}
\includegraphics[scale=0.49]{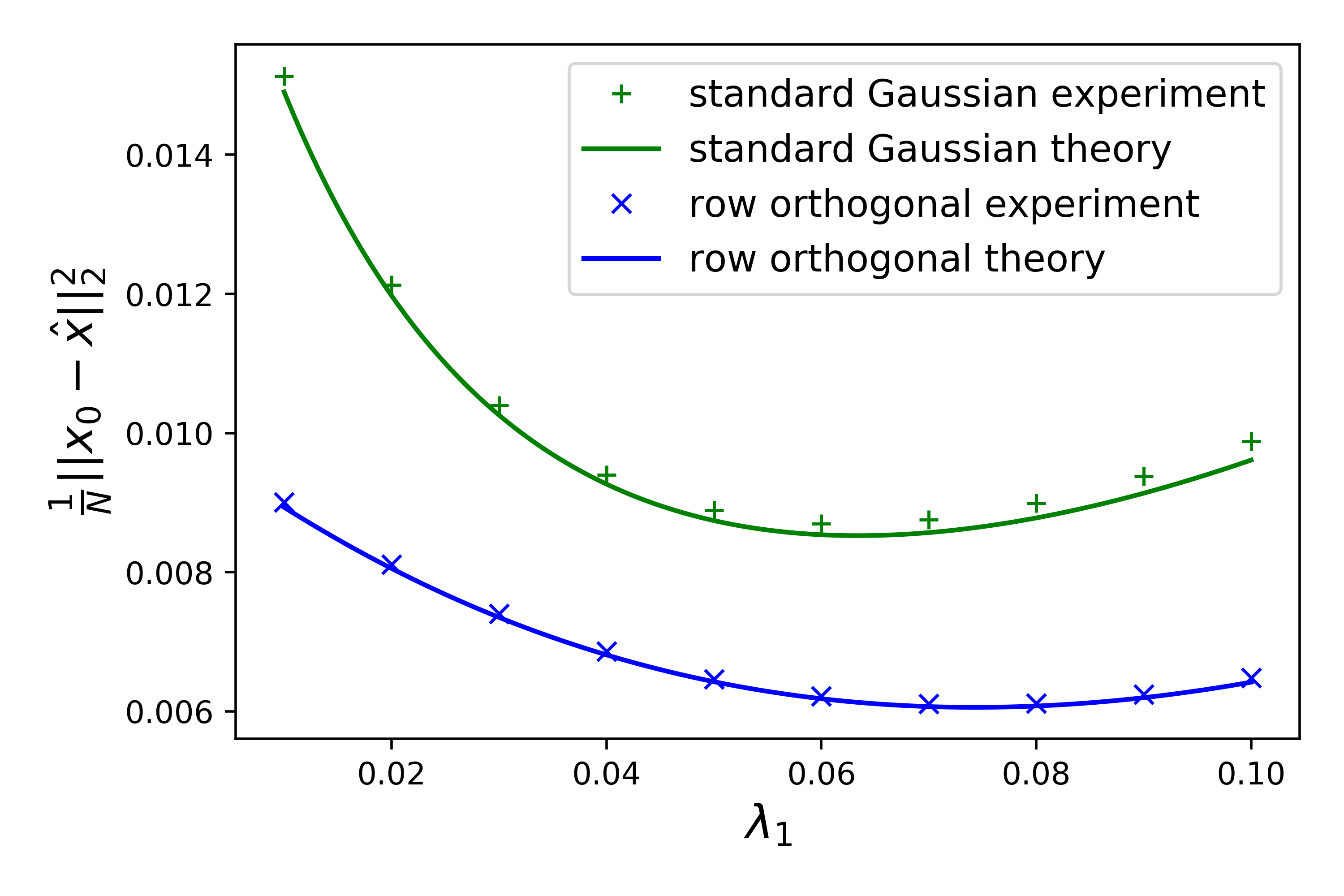}
\includegraphics[scale=0.49]{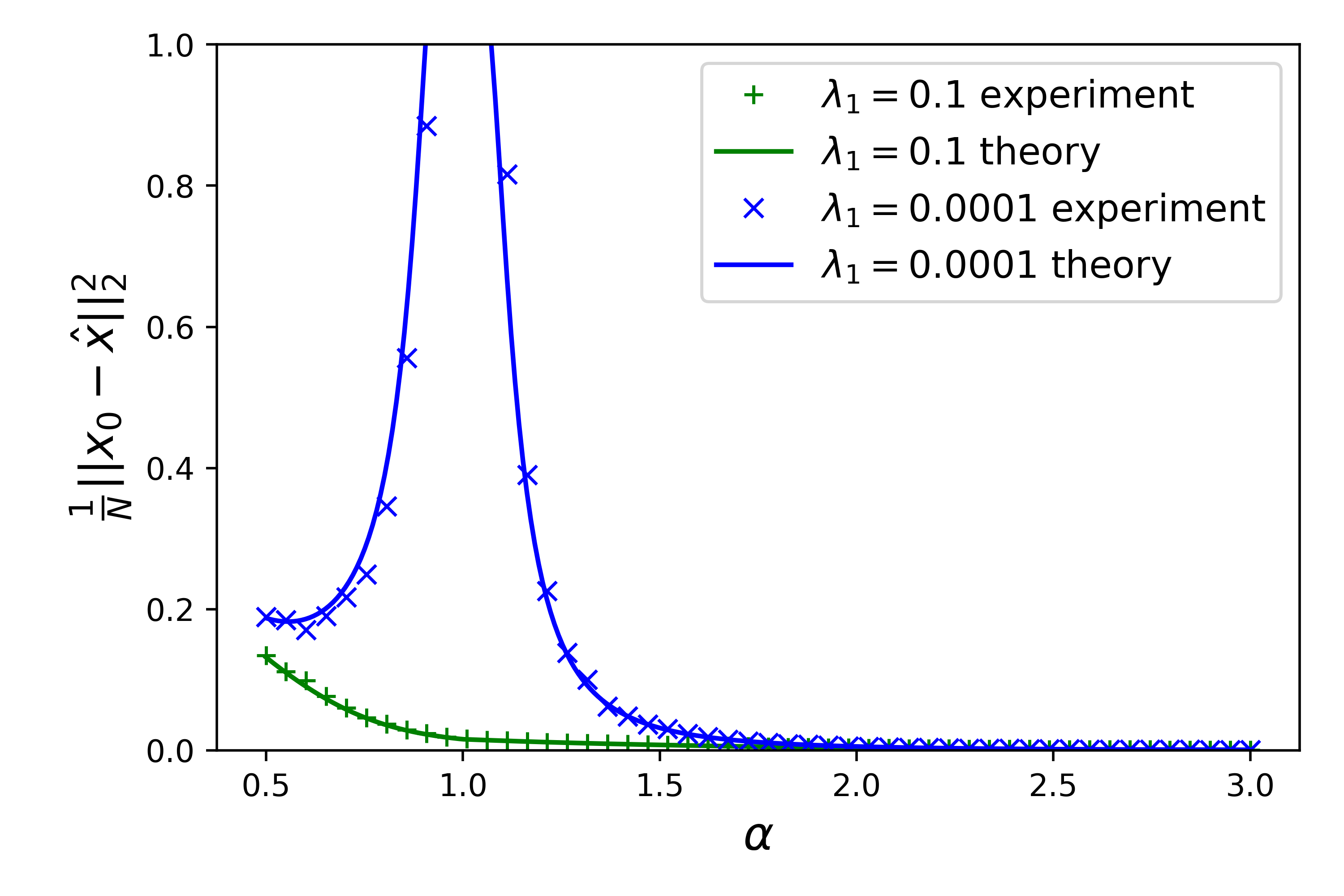}
\caption{Reconstruction error for different realization of model \eqref{main-problem} using the asymptotic formula and numerical simulation with scikit-learn \cite{pedregosa2011scikit}. The  asymptotic predictions are observed to be extremely accurate even at finite low dimension. Left: Reconstruction error for a sparse vector using Gaussian and row-orthogonal matrices with LASSO as regularization function using $M,N = 200,100$.  Right: An example of a double descent \cite{belkin2019reconciling} behavior and its overfitting peak at $\alpha = M/N = 1$ as a function of the sampling ratio. Regularization (here $\ell_1$) explicitly removes the peak (green) to give a smooth curve. The plots were generated using the toolbox from \url{https://github.com/cgerbelo/Oracle_VAMP.}}
\label{default}
\end{center}
\end{figure}

\paragraph{Overparametrization and double descent ---} In the second setup, we consider the effect of the aspect ratio on the reconstruction performance of a sparse vector. We want to reproduce the double descent phenomenon that was observed and discussed recently in several papers \cite{belkin2019reconciling,hastie2019surprises,mitra2019understanding,mei2019generalization} in linear regression (but appeared in some form already in \cite{opper1996statistical}). In order to provide a minimal model of such a phenomenon,  we follow the intuition proposed in \cite{advani2017high}, underlining that the eigenvalue distribution of $\mathbf{C}$ must be divergent (but still be integrable) at $\lambda = 0$ for $\alpha=1$. While the Marchenko-Pastur density is typical of Gaussian data, we can here use random matrices with any spectrum. We choose to sample the singular values of $\mathbf{F}$ from the uniform distribution $\mathcal{U}(\left[(1-\alpha)^{2},(1+\alpha)^{2}\right])$. This leads to the following distribution for the eigenvalues of $\mathbf{C}$:
\begin{equation}
\label{indicator}
    \lambda_{\mathbf{C}} \sim \max(0,1-\alpha)\delta(0)+\min(1,\alpha)\left(\frac{1}{2((1+\alpha)^{2}-(1-\alpha)^{2})}\mathbb{I}_{\{\sqrt{y}\in [(1-\alpha)^{2},(1+\alpha)^{2}]\}}\frac{1}{\sqrt{y}}\right),
\end{equation}
where $\mathbb{I}$ is the indicator function. Our results are shown in Fig. \ref{default} using $M = \lfloor \alpha N \rfloor, N=250$, $\Delta_{0} = 0.05$, for two values of the regularization parameter $\lambda_{1} = 10^{-4},10^{-1}$. Each point is an average over a hundred realizations.
We recover the double descent with the vanishingly small regularization (blue curve). Note that the error peak can be moved to any point $p$ on the x-axis by taking $\mathcal{U}(\left[(p-\alpha)^{2},(p+\alpha)^{2}\right])$. Multiple descents can also be obtained by adding several distributions of the form \eqref{indicator} with different shifts $p$.  Augmenting the regularization to reach a realistic LASSO, however, is found to remove the error peak (green curve). As before, one observed striking agreement between the asymptotics and the simulation. Our formulas generalize here the results of \cite{mitra2019understanding} for any distribution of singular values.

\subsection{Sketch of Proof}

We briefly sketch here the proof techniques that we use. Consider the following modification of problem~\eqref{main-problem}, where an additional $\ell_{2}$ penalty is added to enforce strong convexity and the potentially non-differentiable penalty function $f$ is replaced by its twice differentiable approximation $\tilde{f}$, using for example \cite{azagra2007smooth}:

\begin{equation}
    \mathbf{x^*_{\lambda_2}} = \argmin_{\mathbf{x}\in\mathbb{R}^{N}} \frac{1}{2} \norm{\mathbf{y}-\mathbf{F}\mathbf{x}}_{2}^{2}+\tilde{f}(\mathbf{x})+\frac{\lambda_{2}}{2}\norm{\mathbf{x}}_{2}^{2}.\label{mod_problem}
\end{equation}

The main idea behind the proof is a variation of the approach pioneered in \cite{bayati2011dynamics}, and leverages on the results of \cite{rangan2019vector} who computed the errors obtained by vector approximate message-passing (VAMP) on this problem. The sequence of steps behind the proof are the following: (i) first we show that for each instance of problem~\eqref{mod_problem} with large enough $\lambda_{2}$, a sequence of iterates of VAMP that converges towards the solution of~\eqref{mod_problem} can be found. To do this, we use a modified version of VAMP, that we call oracle-VAMP. (ii) the statistical properties of these iterates can be analytically tracked (in the asymptotic high-dimensional limit) by exact equations with a fixed point that yields the same result as the replica equations presented in~\thref{theorem1}. This shows that for large enough $\lambda_{2}$, the replica prediction is correct. (iii) An analytic continuation theorem on the parameter $\lambda_{2}$ is then used to extend the result obtained on problem~\eqref{mod_problem} to problem~\eqref{main-problem}.

\section{Vector approximate message-passing (VAMP)}
We briefly present the VAMP algorithm \cite{rangan2019vector} along with its properties and oracle version that will be fundamental for our proof, which builds on results established mostly in \cite{rangan2019vector} and \cite{fletcher2016expectation}.  VAMP is also linked with the expectation-propagation strategy \cite{minka2001family}, as well as other algorithms  \cite{cakmak2014s,ma2017orthogonal}. \cite{rangan2019vector}, however,
has the significant trait to provide rigourously derived state evolution equations

\paragraph{Maximum a posteriori formulation of VAMP ---} In a Bayesian framework, problem \eqref{main-problem} corresponds to a maximum a posteriori (MAP) problem.
The MAP formulation of vector approximate message-passing for \eqref{main-problem} reads:
\begin{subequations}
\label{VAMP_alg}
\begin{align}
    &\mbox{\emph{Choose initial $A_{10}$ and isotropically distributed $\mathbf{B}_{10}$}} \notag \\
    \hat{\mathbf{x}}_{1k} &= \mbox{$\prox_{\frac{1}{A_{1k}}f}$}\left(\frac{\mathbf{B}_{1k}}{A_{1k}}\right) \quad \quad \hat{\mathbf{x}}_{2k} = (\mathbf{F}^{T}\mathbf{F}+A_{2k}\rm{Id})^{-1}(\mathbf{F}^{T}y+\mathbf{B}_{2k}) \\
    V_{1k} &= \frac{\langle \mbox{$\prox'_{\frac{1}{A_{1k}}f}$} \rangle}{A_{1k}} \hspace{1.8cm} V_{2k} = \frac{1}{N}\mbox{Tr}\left[(\mathbf{F}^{T}\mathbf{F}+A_{2k}\rm{Id})^{-1}\right] \\
    A_{2k} &= \frac{1}{V_{1k}}-A_{1k}\hspace{2.05cm} A_{1,k+1} = \frac{1}{V_{2k}}-A_{2k} \\
    \mathbf{B}_{2k} &= \frac{\hat{\mathbf{x}}_{1k}}{V_{1k}}-\mathbf{B}_{1k} \hspace{2cm} \mathbf{B}_{1,k+1} = \frac{\hat{\mathbf{x}}_{2k}}{V_{2k}}-\mathbf{B}_{2k}
\end{align}
\end{subequations}
where $\langle \cdot \rangle$ is an element-wise averaging operator $\langle \mathbf{x} \rangle = \frac{1}{N}\sum_{i=1}^{N}x_{i}$, and the vector valued proximal operator is defined as :
\begin{equation}
    \forall \gamma \in \mathbb{R}^{+}, \mathbf{x}, \mathbf{y} \in \mathcal{X} \quad  \mbox{$\prox_{\gamma f}$}(\mathbf{y}) \equiv \argmin_{\mathbf{x}} \left\lbrace f(\mathbf{x})+\frac{1}{2\gamma}\norm{\mathbf{x}-\mathbf{y}} _{2}^{2} \right\rbrace.
\end{equation}
It is akin to a projection on the level sets of $f$ tuned with the parameter $\gamma$ and can be evaluated even when the objective function is non-differentiable. Proximal descent methods enjoy a long lasting success in machine learning and signal processing \cite{combettes2011proximal} because of their stability, simplicity to implement and solid theoretical anchoring, notably from a monotone operator theory point of view \cite{bauschke2011convex}. A popular algorithm for solving composite convex optimization problems of the form $\argmin_{\mathbf{x}} \{ f(\mathbf{x})+g(\mathbf{x}) \} $ is the Douglas-Rachford splitting method \cite{parikh2014proximal}, which roughly amounts to successively applying the proximal of $f$ and the one of $g$. It is shown in \cite{fletcher2016expectation}, a connection pursued in \cite{manoel2018approximate}, that VAMP is akin to a Douglas-Rachford descent with parameters that adapt to the local curvature of the cost function.
\par
The state evolution equations of VAMP give the statistical distribution of the iterates in \eqref{VAMP_alg}. The quantities $\mathbf{B_{1}}$ and $\mathbf{B_{2}}$ behave as noisy Gaussian estimates of $\mathbf{x}_{0}$:
\begin{align}
    \mathbf{B}_{1k} &= A_{1k}(\mathbf{x}_{0}+\mathbf{P}_{1k}) \quad \quad \mathbf{B}_{2k} = A_{2k}( \mathbf{x}_{0}+\mathbf{P}_{2k}), \\
    \mbox{where} \quad P_{1k} &\sim \mathcal{N}(0, \tau_{1k}) \hspace{1.7cm} P_{2k} \sim \mathcal{N}(0, \tau_{2k})
\end{align}
with $\tau_{1k}$ and $\tau_{2k}$ the variances of the estimates. The other key quantities are the variances $V_{1k}$ and $V_{2k}$ of the estimates $\hat{\mathbf{x}}_{1k}$ and $\hat{\mathbf{x}}_{2k}$, and their respective MSEs $\mathcal{E}_{1}$ and $\mathcal{E}_{2}$ from \thref{theorem2}, which can also be written (appendix C from \cite{rangan2019vector}):
\begin{equation}
    \label{equation : e2}\mathcal{E}_{2} = \lim_{N \to \infty} \frac{1}{N} \mathbb{E}_{x_{0},P_{2k}}\left[\norm{(\mathbf{F}^{T}\mathbf{F}+A_{2k}\rm{Id})^{-1}(\mathbf{F}^{T}y+\mathbf{B}_{2k})-\mathbf{x}_{0}}^{2}\right]
\end{equation}
The thresholds $A_{1k}$ and $A_{2k}$
of the proximal operators are adjusted to the variance of the noisy estimates of the teacher vector. The full state evolution (SE) equations are then given by \eqref{VAMP_se}. \\
\par
These equations can be solved analytically if the teacher distribution $\phi(x_{0})$ is known. In practice, all the averages are empirical and $\tau_{10}$ is initialized with the empirical variance of $\mathbf{B}_{10}$. Note that the SE equations only hold if VAMP is initialized with an isotropically distributed vector $\mathbf{B}_{10}$ which empirically converges with second order moment. Three additional assumptions on the denoiser function are required for the state evolution theorem in \cite{rangan2019vector} to hold. These are automatically verified in the convex MAP case, as properties of the proximal mapping. This is reminded in appendix \ref{appendix:analysis_framework}. The next lemma characterizes the fixed point of \eqref{VAMP_alg}.
\begin{lemma}
\label{lemma3}
For any pair $(A_{1},A_{2})$ such that $A_{1}+A_{2} = \frac{1}{V}$, the fixed point of iterations \eqref{VAMP_alg} solves the minimization problem \eqref{main-problem}.
\end{lemma}

\emph{Proof}. See appendix \ref{appendix : prox_fixed_point}.

\paragraph{From VAMP to oracle-VAMP ---} We remind the reader that our goal is to find a convergent sequence of VAMP iterates that obey the state evolution equations, and reaches a fixed point solving \eqref{main-problem}. To simplify the analysis, we want to choose fixed values for the variance parameters of iterations \eqref{VAMP_alg}, i.e. $A_1$, $A_2$, $V_1$ and $V_2$. To do so, we initialize $A_1$ in its value given by the fixed point of the state evolution equations. Taking $\mathbf{B_{10}}$ an isotropically distributed vector, VAMP will then yield iterates that obey the SE equations while keeping constant variance parameters. Equations~\eqref{VAMP_alg} turn into a simpler algorithm, that we call oracle-VAMP:
\begin{subequations}
\label{dr-vamp}
\begin{align}
    \hat{\mathbf{x}}_{1k} &= \mbox{$\prox_{\frac{1}{A_{1}}f}$}\left(\frac{\mathbf{B}_{1k}}{A_{1}}\right) \quad \quad \hat{\mathbf{x}}_{2k} = \mbox{$\prox_{\frac{1}{2A_{2}}\vert \vert y-Fx \vert \vert _{2}^{2}}$} \left(\frac{\mathbf{B}_{2k}}{A_{2}}\right) \\
    \mathbf{B}_{2k} &= \frac{\hat{\mathbf{x}}_{1k}}{V_{1}}-\mathbf{B}_{1k} \hspace{1.85cm} \mathbf{B}_{1,k+1} = \frac{\hat{\mathbf{x}}_{2k}}{V_{2}}-\mathbf{B}_{2k} ,
\end{align}
\end{subequations}
where the coefficients $A_{1}$ and $A_{2}$ verify:
\begin{align}
\label{se_fixed_point}
    V = \mathcal{S}_{\mathbf{F}^T \mathbf{F}}(-A_{2}),\quad \quad A_{1}+A_{2} = \frac{1}{V}
\end{align}
with $\mathcal{S}_{\mathbf{C}}$ the Stieltjes transform with respect to the spectral measure of $\mathbf{C}$. \newline
\eqref{dr-vamp} can then be rewritten as a single iteration on $\mathbf{B_2}$:
\begin{align}
\label{oo}
    \mathbf{B}_{2,k+1} &= \mathcal{O}_{1} \circ \mathcal{O}_{2}(\mathbf{B}_{2k}) \\
    \mbox{where} \quad \mathcal{O}_{1} &= \frac{1}{V}\mbox{$\prox_{\frac{1}{A_{1}}f}$}(\frac{.}{A_{1}})-\rm{Id}, \quad \mbox{and} \quad  \mathcal{O}_{2} = \left(\frac{1}{V}\mbox{$\prox_{\frac{1}{2A_{2}}\vert \vert \mathbf{y}-\mathbf{F}\mathbf{x} \vert \vert _{2}^{2}}$}(\frac{.}{A_{2}})-\rm{Id}\right).
\end{align}
which becomes the Peaceman-Rachford operator \cite{peaceman1955numerical} if $A_{1} = A_{2} = \frac{1}{2V}$ is artificially prescribed (note that such a manipulation renders the state evolution equations invalid). 

We point out that, for an $\ell_2$-penalty, oracle-VAMP iterations reduce to a one step process which is identical to a ridge regression with parameter $A_{2}$ (see appendix \ref{appendix : cvx_opt}). To generalize this result to any convex regularization, we derive upper bounds for the Lipschitz constants of the iteration. 
\paragraph{Lipschitz constants of oracle-VAMP iteration ---}
To characterize oracle-VAMP iterations, we will make use of convenient proximal operator properties, such as firm nonexpansiveness: for all $\mathbf{x},\mathbf{y}$ in the input space, the following inequality holds
\begin{equation}
    \langle \mathbf{x}-\mathbf{y},\mbox{$\prox_{f}$}(\mathbf{x})-\mbox{$\prox_{f}$}(\mathbf{y}) \rangle \geqslant \norm{\mbox{$\prox_{f}$}(\mathbf{x})-\mbox{$\prox_{f}$}(\mathbf{y})}^{2}.
\end{equation}
We remind a few useful definitions from convex analysis in the appendix \ref{appendix : prox_prop}. Moreover, we will use the following result from \cite{giselsson2016linear}:
\begin{proposition}(Proposition 2 from \cite{giselsson2016linear})
\label{proposition : gisel_idh}
Assume that $f$ is $\sigma$-strongly convex and 
    $\beta$-smooth and that $\gamma \in ]0, \infty[$. Then $\prox_{\gamma f} - \frac{1}{1+\gamma \beta}\rm{Id}$ is $\frac{1}{\frac{1}{1+\gamma \beta}-\frac{1}{1+\gamma \sigma}}$-cocoercive
    if $\beta > \sigma$ and 0-Lipschitz if $\beta = \sigma$.
\end{proposition}
Let $(\sigma_{1},\beta_{1})$ be the strong convexity and smoothness
constants of the regularization function. Let $(\sigma_{2},\beta_{2})$ be the corresponding constants of the squared loss $(\mathbf{x} \mapsto \frac{1}{2} \norm{\mathbf{y}- \mathbf{F x}}_2^2)$. We easily find $\sigma_{2} = \lambda_{min}(\mathbf{C})$ and $\beta_{2} = \lambda_{max}(\mathbf{C})$, the minimal and maximal eigenvalue of $\mathbf{C}$. 
Using these results and the properties of the fixed point of the state evolution equations, we get the following upper bounds on the Lipschitz constant of the iteration \eqref{oo}, depending on the aspect ratio ${\alpha = M/N}$ and constants $(\sigma_{1,2},\beta_{1,2})$.

\begin{lemma}(Lipschitz constants of iteration~\eqref{oo}) 
\label{lipschitz}
\paragraph{Lipschitz constant of $\mathcal{O}_{1}$ ---}
The Lipschitz constant $\mathcal{L}_1$ of the operator $\mathcal{O}_{1}$ in the cases ${0<\sigma_{1}<\beta_{1}}$, ${0<\sigma_{1}=\beta_{1},0=\sigma_{1}=\beta_{1}}$ respectively reads:
\begin{equation}
    \mathcal{L}_1 = \max \left(\frac{\abs{A_{2}-\sigma_{1}}}{A_{1}+\sigma_{1}},\frac{\abs{\beta_{1}-A_{2}}}{A_{1}+\beta_{1}}\right) \thickspace , \thickspace
    \mathcal{L}_1 =\sqrt{\left(\frac{(A_{2}^{2}-A_{1}^{2})}{(A_{1}+\sigma_{1})^{2}}+1\right)} \thickspace , \thickspace
    \mathcal{L}_1 =\max\left(1,\frac{{A}_{1}}{A_{2}}\right). \label{Lipschitz_o1}
\end{equation}

\paragraph{Lipschitz constant of $\mathcal{O}_{2}$ ---}
The Lipschitz constant $\mathcal{L}_2$ of the operator $\mathcal{O}_{2}$ reads

\begin{equation}
  \mathcal{L}_2=\max \left(\frac{\abs{A_{1}-\lambda_{min}(\mathbf{F}^T \mathbf{F})}}{A_{2}+\lambda_{min}(\mathbf{F}^T \mathbf{F})},\frac{\abs{\lambda_{max}(\mathbf{F}^T \mathbf{F})-A_{1}}}{A_{2}+\lambda_{max}(\mathbf{F}^T \mathbf{F})}\right). \label{lipschitz_o2}
\end{equation}
\end{lemma}

\emph{Proof}. See appendix \ref{appendix : cvx_opt}. \\
\par
The case $0<\sigma_{1}<\beta_{1},\alpha>1$ yields the same constant as the 
one derived in \cite{fletcher2016expectation}, which studies a more general version of VAMP, and where the proof method relies on the analysis of the Jacobian of the prescribed iteration. Note that all those constants reduce to 1, i.e. to non-expansive operators if $A_{1}=A_{2}$ is set, which is consistent with the non-expansiveness of the Peaceman-Rachford operator \cite{peaceman1955numerical}.

\section{Proof of \thref{theorem2}}

To prove \thref{theorem2}, we will theoretically solve problem~\eqref{mod_problem} with the oracle-VAMP~\eqref{dr-vamp} algorithm. In addition to its SE equations, it enjoys the following useful property: the fixed point of oracle-VAMP solves the minimization problem we are interested in, i.e. problems~\eqref{main-problem} or \eqref{mod_problem} depending on the choice of penalty function. This is proven in appendix \ref{appendix : prox_fixed_point}.

Knowing those two points, it is now tempting -- but incorrect -- to conclude the proof here by stating that the desired MAP estimator's properties are described by SE equations since it trivially belongs to the trajectory of VAMP initialized at its solution. However this reasoning is flawed : for the SE equations to hold, we need to initialize oracle-VAMP with an isotropically distributed vector. Having no such information on the solution, we need to find at least one converging trajectory starting from an isotropically distriubted vector.

\paragraph{Convergence of oracle-VAMP sequence ---} The convergence analysis of oracle-VAMP is based on its similarities with the Douglas-Rachford algorithm . We have derived in Lemma~\ref{lipschitz} upper bounds on the Lipschitz constant of the oracle-VAMP iterations. We now turn to problem \eqref{mod_problem} and start by proving that, for sufficiently large regularization parameter $\lambda_2$, the operator $\mathcal{O}_1 \circ \mathcal{O}_2$ in iteration~\eqref{oo} is a contraction. We first state two useful lemmas that will help in bounding the Lipschitz constants.

\begin{lemma}
\label{stieltjes-lemma}
At the fixed point of the state evolution equations, the coefficients $A_{1}$ and $A_{2}$ verify:
\begin{equation}
    V = \mathcal{S}_{\mathbf{C}}(-A_{2}), \quad  \quad V = \mathcal{S}_{\mathcal{H}_{f}(\hat{\mathbf{x}})}(-A_{1})
\end{equation}
where $\mathcal{H}_{f}(\hat{\mathbf{x}})$ is the Hessian of the penalty function taken at the fixed point of the algorithm.

\end{lemma}

\emph{Proof.}  See appendix \ref{appendix : stieltjes-lemma-proof}.\\

A direct consequence of Lemma \ref{stieltjes-lemma} is to give upper and lower bounds on $A_{1}$ and $A_{2}$, which are the constituents of the upper bounds on the Lipschitz constants of iteration \eqref{oo}.

\begin{lemma}
\label{lemma-bounds}
 At the fixed point of the state evolution equations, we have:
 \begin{equation}
     \lambda_{min}(\mathbf{F}^{T}\mathbf{F}) \leqslant A_{1} \leqslant  \lambda_{max}(\mathbf{F}^{T}\mathbf{F}), \quad \quad \sigma_{1} \leqslant A_{2} \leqslant \beta_{1}. \label{ineq}
 \end{equation}
\end{lemma}

\emph{Proof.} See appendix \ref{appendix : proof-lemma-bounds}.
\\

We now consider problem \eqref{mod_problem}, which involves $h = \tilde{f} + \frac{\lambda_2}{2} \norm{x}_2^2$ as regularization function. Let $(\tilde{\sigma}_{1},  \tilde{\beta}_{1})$ be the strong convexity and smoothness constants of $\tilde{f}$. Using the second-order definition of strong convexity and smoothness, it is straightforward to obtain $\sigma_{1} = \tilde{\sigma}_{1} +\lambda_{2}, \beta_{1} = \tilde{\beta}_{1} +\lambda_{2}$. By arbitrarily increasing $\lambda_2$, we can thus accordingly augment $\sigma_{1},  \beta_{1}$. Using Lemma \ref{lemma-bounds}, we see that $A_{2}$ grows with $\lambda_2$, while $A_1$ remains bounded.\\

We can easily see that the Lipschitz constant of operator $\mathcal{O}_1$ given in~\eqref{Lipschitz_o1} is bounded by a constant $C$. We focus on the Lipschitz constant~\eqref{lipschitz_o2} of $\mathcal{O}_2$. Using inequalities~\eqref{ineq}, we get
\begin{equation}
    \mathcal{L}_2 \leqslant \dfrac{\lambda_{max}(\mathbf{F}^T \mathbf{F}) - \lambda_{min}(\mathbf{F}^T \mathbf{F})}{\tilde{\sigma}_1 + \lambda_2 + \lambda_{min}(\mathbf{F}^T \mathbf{F})}.
\end{equation}
We choose $\lambda_2$ large enough, for instance 
\begin{equation}
    {\lambda_2 > C(\lambda_{max}(\mathbf{F}^T \mathbf{F}) - \lambda_{min}(\mathbf{F}^T \mathbf{F})) - \tilde{\sigma}_1 - \lambda_{min}(\mathbf{F}^T \mathbf{F})}
\end{equation},  which induces
\begin{equation}
      \mathcal{L}_2 \leqslant \dfrac{\lambda_{max}(\mathbf{F}^T \mathbf{F}) - \lambda_{min}(\mathbf{F}^T \mathbf{F})}{\tilde{\sigma}_1 + \lambda_2 + \lambda_{min}(\mathbf{F}^T \mathbf{F})} < 1.
\end{equation}
Therefore the Lipschitz constant $\mathcal{L}$ of $\mathcal{O}_{1} \circ \mathcal{O}_{2}$ verifies
\begin{equation}
\mathcal{L} \leqslant \mathcal{L}_1 \mathcal{L}_2 < 1.
\end{equation}
For problem~\eqref{mod_problem} with sufficiently large $\lambda_2$, iteration \eqref{oo} becomes a contraction and we force the convergence. We provide plots in appendix \ref{appendix : e_net} to illustrate this claim on the elastic net problem. Any oracle-VAMP iterates sequence will thus be convergent in this setting. This sequence can be properly initialized and will be described by state evolution equations. According to Lemma~\ref{lemma3}, the estimator returned by oracle-VAMP does solve our modified minimization problem. Moreover, the MSE and variance of this estimator obey the state evolution fixed point equations: this concludes the proof for \thref{theorem2}, for high enough $\lambda_2$. \thref{theorem2} is equivalent to \thref{theorem1},  as shown in \ref{appendix : rep_eq_se}.

\paragraph{Analytic continuation ---} We now need to continuate the result for any $\lambda_{2}$, which is only possible because of the convexity of the problem and its analyticity. We first invoke the optimality condition on the convex problem \eqref{mod_problem} which gives the following prescription for the solution:
\begin{equation}
    \left( \mathbf{F}^{T}\mathbf{F}+\lambda_{2}\rm{Id}+\nabla f \right)\mathbf{x} = \mathbf{F}^{T}\mathbf{y}.
\end{equation}
Using the analytic inverse function theorem \cite{krantz2002primer}, this clearly prescribes an analytic solution for $\mathbf{x}$ in $\lambda_{2}$. We then turn to the SE equations~\eqref{VAMP_se}, which can also be written as~\eqref{main-result} using the replica formalism, as highlighted in \ref{appendix : rep_eq_se}. Appendix \ref{appendix : prox_prop2} helps isolate the additional ridge contribution, and shows that equations~\eqref{main-result} are analytic in $\lambda_2$. The implicit function theorem \cite{krantz2012implicit} ensures that the scalar quantities defined by the equations, including the mean squared error, are analytic in $\lambda_{2}$. We can conclude using the analytic continuation property \cite{krantz2002primer} that the replica formula and all the SE quantities hold true whatever the value of $\lambda_{2}$. In particular, taking $\lambda_2 = 0$ provides the MSE of the modified problem which only differs of the original problem \eqref{main-problem} by the use of a twice differentiable penalty function $\tilde{f}$. Going from the differentiable relaxation to the real problem only relies on finding an appropriate sequence of twice differentiable functions $(f_{n})_{n \in \mathbb{N}}$ converging towards $f$ and taking the limit $n \rightarrow \infty$ inside well-defined scalar quantities. It will not be detailed here as it remains intuitive: non-differentiable $f$ in \eqref{main-problem} can be naturally approximated by appropriate sequences of differentiable functions, see for example the remark in \cite{el2013robust}.
Figure \ref{default} indeed shows that the prediction holds for a plain LASSO, with no additional ridge or differentiable approximation. Although all scalar quantities can be continuated, we have no theorem for the continuation of the Gaussian property. We conjecture it to be true for all $\lambda_{2}$ in the second part of \thref{theorem1}. \\
\quad \\
\section{A note on non-separable denoisers}

In a recent paper \cite{fletcher2018plug}, the state evolution analysis of VAMP is extended to a large class of non-separable convex denoisers which verify the so-called \emph{convergence under Gaussian noise} property, building upon previous work on convex, non-separable regularization in message-passing algorithms in \cite{berthier2017state}. This broader class includes the following operations : group-based denoisers, convolutional ones and neural nets as well as singular value-thresholding. The state evolution equations are thus valid for this family of denoisers. Additionally, the Lipschitz constants prescribed by Lemma \ref{lipschitz} still hold for non-separable denoisers, as the proof only depends on the strong-convexity and smoothness assumptions. According to \cite{fletcher2018plug}, the variance terms for the non-separable case $A_{1},A_{2}$ are defined according to :
\begin{equation}
    A_{i} = \frac{1}{N}\sum_{n=1}^{N}\frac{\partial g_{i,n}(\mathbf{r},\gamma)}{\partial r_{i,n}}
\end{equation}

where $\mathbf{g}_{i} : \mathbb{R}^{N}\to \mathbb{R}^{N}$ is the proximal of the loss function for $A_{2}$ and of the regularization for $A_{1}$, and $g_{n}$ its n-th component. This is exactly the normalized trace of the Jacobian matrix of the proximal, and encompasses the element-wise averaging operator defined in \ref{VAMP_alg} for the separable case. Using the expression prescribed by appendix \ref{appendix : prox_prop}, we see that Lemmas \ref{stieltjes-lemma} and \ref{lemma-bounds} still hold. We can thus enforce convergence of any VAMP trajectory and complete the proof in the same way it was done for separable denoisers.\\
\quad \\
The scripts used to generate all plots can be found at \url{https://github.com/cgerbelo/Oracle_VAMP}.

\renewcommand{\abstractname}{Acknowledgements}
\begin{abstract}We thank Marc Lelarge, Antoine Maillard,  Sundeep Rangan and Lenka Zdeborov\'a for insightful discussions. This work benefited from state aid managed by the Agence Nationale de la Recherche under the "Investissements d’avenir" program with the reference ANR-19-P3IA-0001, and from the ANR PAIL. We also acknowledge support from the chaire CFM-ENS "Science des donn\'ees".
\end{abstract}
\newpage

\bibliographystyle{alpha}
\bibliography{main}
\newpage
\appendix

\section{Heuristic replica derivation from statistical physics}
\label{appendix : replica_full}
For completeness, we give here the heuristic derivation from the replica method. To characterize the MAP estimator properties, it is useful to compute the posterior distribution normalization factor $\mathcal{Z}(\mathbf{y},\mathbf{F})$. In a physics perspective, $\mathcal{Z}$ is the partition function of the problem, and we can define the corresponding free energy averaged on the size of the signal $\Phi = \frac{1}{N} \log \mathcal{Z}$. This quantity is known to be self-averaging: when $N$ goes to infinity, $\Phi$ concentrates on its average with respect to the distribution of data matrix elements. Hence we will focus on the averaged free energy $\frac{1}{N}\mathbb{E}_{\mathbf{F},\mathbf{x}_{0}}(\log \mathcal{Z})$. However, directly computing the average of the logarithm of $\mathcal{Z}$ is analytically intractable. We will replace this computation by an easier integral following the so-called replica trick:
\begin{equation}
    \lim_{N \rightarrow \infty} \frac{1}{N}\mathbb{E}_{\mathbf{F},\mathbf{x}_{0},\mathbf{w}}(\log \mathcal{Z}) =  \lim_{N \rightarrow \infty} \frac{1}{N} \lim_{n \to 0} \frac{\mathbb{E}_{\mathbf{F},\mathbf{x}_{0},\mathbf{w}}(\mathcal{Z}^{n})-1}{n}.
\end{equation}
The partition function reads
\begin{equation}
    \mathcal{Z}(\mathbf{y},\mathbf{F},\mathbf{w}) = \int \prod_{i=1}^{N}dx_{i}\prod_{i=1}^{N}p(x_{i})\prod_{\mu=1}^{M}\frac{1}{\sqrt{2\pi \Delta}}e^{-\frac{1}{2\Delta}\left(y_{\mu}-\sum_{i=1}^{N}F_{\mu i}x_{i}\right)^{2}} .
\end{equation}
Introducing n replicas of the system, we now consider
\begin{equation}
\mathbb{E}_{\mathbf{F},\mathbf{x}_{0},\mathbf{w}}(Z^{n}) = \int\prod_{i,a}dx_{i}^{a}\prod_{i,a}p(x_{i}^{a})\prod_{\mu}\mathbb{E}_{\mathbf{F},\mathbf{x}_{0},\mathbf{w}}\frac{1}{\sqrt{2\pi \Delta}}e^{-\frac{1}{2\Delta}\sum_{a=1}^{n}\left(\sum_{i=1}^{N}F_{\mu i}x_{0,i}+w_{\mu}-\sum_{i=1}^{N}F_{\mu i}x_{i}^{a}\right)^{2}}.\label{Z^n}
\end{equation}
Following the statistical physics tradition, the computation goes on by introducing the following order parameters:
\begin{align}
    m^{a} &= \frac{1}{N}\sum_{i=1}^{N}x_{i}^{a}x_{0,i} \quad a = 1,2,... n \\
    Q^{a} &= \frac{1}{N}\sum_{i=1}^{N}(x_{i}^{a})^{2} \quad a = 1,2, ... n \\
    q^{ab} &= \frac{1}{N}\sum_{i=1}^{N}x_{i}^{a}x_{i}^{b}  \quad a = 1,2, ... n. 
    \label{order_parameters}
\end{align}
The parameters $m$ and $q$ respectively quantify the overlap between the teacher and the student weights, and the overlap between the replicas and the student weights. $Q$ is a norm-like parameter for the replicas of the student weights. Those order parameters carry physical meaning and will eventually provide information about the MAP estimator. A key step of the computation is to carry out the average on the matrix elements, which depends on the statistics assumed for the data matrix. In the case of i.i.d. elements, this can be done using the central limit theorem, see~\cite{krzakala2012probabilistic}.  An extension from the i.i.d. case has been proposed both in the context of statistical physics \cite{parisi1995mean} and signal processing \cite{kabashima2014signal}. In that case, $\mathbf{F}$ is considered rotationally invariant and $\mathbf{C} = \mathbf{F}^T \mathbf{F}$ has an arbitrary and well-defined singular value distribution $\mu(\lambda)$ with compact support. We can define its minimum $\lambda_{min}$ and its maximum $\lambda_{max}$. Let us recall useful transform definitions. The Stieltjes transform associated to $\mu(\lambda)$ is
\begin{equation}
    \mathcal{S}_{\mathbf{C}}(z) = \int_{\lambda_{min}}^{\lambda_{max}} \dfrac{d\lambda \mu(\lambda)}{\lambda-z}= \mathbb{E} \left[ \dfrac{1}{\lambda-z}\right]
\end{equation}
and is correctly defined outside of $\mu$'s support. The corresponding R-transform is 
\begin{equation}
    \mathcal{R}_{\mathbf{C}}(x) = \mathcal{S}_{\mathbf{C}}^{-1}(-x) - \frac{1}{x}. 
\end{equation}
Throughout this paper, the support of the considered matrices is always comprised in $\mathbb{R}_+$. The Stieltjes transform is hence well-defined on strictly negative values, and the R-transform is defined in a neighborhood of $0$, included in $\mathbb{R}_-^*$.
We first perform the averages over $\mathbf{x}_{0}$ and $\mathbf{w}$ in~\eqref{Z^n}, we can then average on $\mathbf{F}$ using the asymptotic form of the Harish-Chandra-Itzykson-Zuber integral~\cite{Tanaka_2008}. Since $\mathbf{F}$ is rotationally invariant, for any function $\phi$ of $\mathbf{F}$:
\begin{equation}
    \mathbb{E}_{\mathbf{F}}\left[ \phi(\mathbf{F})\right] = \mathbb{E}_{\mathbf{F}}\left[ \mathcal{D}\mathbf{U} \hspace{.1cm}\mathcal{D} \mathbf{V} \hspace{.1cm}\phi (\mathbf{UFV}^T) \right]
\end{equation}
where integrating on $\mathcal{D} \mathbf{U}$, $\mathcal{D} \mathbf{V}$ represents averages over the ensemble of orthogonal matrices using the Haar measure. The Harish-Chandra-Itzykson-Zuber integral then allows to write the result as a function that depends only on the singular value asymptotic distribution $\mu(\lambda).$  This step can only be done when the matrix $\mathbf{F}$ is rotationally invariant, which is an assumption of \thref{theorem1} and \thref{theorem2}.\\

Finally, we take the so-called replica symmetric ansatz which assumes that order parameters are the same for every replica. We can thus remove the subscripts in~\eqref{order_parameters} and only keep three order parameters and their conjugate parameters. The replicated partition function reads
\begin{equation}
    \mathbb{E}_{\mathbf{F},\mathbf{x}_{0},\mathbf{w}}(Z^{n}) = \int dQ \hspace{.1cm} d\hat{Q} \hspace{.1cm}dq \hspace{.1cm} d\hat{q} \hspace{.1cm} dm \hspace{.1cm} d\hat{m}\hspace{.1cm}e^{n N \Phi(Q, q, m, \hat{Q}, \hat{q}, \hat{m})}.
\end{equation}
Assuming that we can take the limit $N \rightarrow \infty$ before the limit $n \rightarrow 0$, the integral will concentrate on its saddle-point
\begin{equation}
\mathbb{E}_{\mathbf{F},\mathbf{x}_{0},\mathbf{w}}(Z^{n}) = e^{n N \Phi^*}.
\end{equation}
The final form of the average free energy is given in~\cite{kabashima2014signal} and reads
\begin{multline}
    \Phi(Q,q,m,\hat{Q},\hat{q},\hat{m}) = \mathcal{G}_{\mathbf{C}}\left(-\dfrac{Q-q}{\Delta}\right) +\left( - \dfrac{\mathbb{E}(x_0^2) - 2 m + q}{\Delta} + \dfrac{\Delta_0 (Q - q)}{\Delta^2}\right) \mathcal{G}_\mathbf{C}'\left( - \dfrac{Q - q}{\Delta}\right) \\ + \dfrac{Q \hat{Q}}{2} - m \hat{m} + \dfrac{q \hat{q}}{2}
    + \int dx_0\hspace{.1cm}\phi(x_0) \int dz \dfrac{e^{-\frac{z^2}{2}}}{\sqrt{2 \pi}} \log \left\lbrace \int dx \hspace{.1cm} e^{-\frac{1}{\Delta}f(x)-\frac{\hat{Q}+\hat{q}}{2}x^2 + \hat{m} x x_0 + z \sqrt{\hat{q}}x} \right\rbrace \label{free_energy_RI}
\end{multline}
where $\mathcal{G}_\mathbf{C}$ appears when performing the Harish-Chandra-Itzykson-Zuber integral, and is defined with respect to $\mu(\lambda)$ as
\begin{equation}
    \mathcal{G}_\mathbf{C}(x)= \dfrac{1}{2}\text{Sup}_{\Lambda} \left\lbrace - \int d \lambda \mu(\lambda) \log |\Lambda - \lambda| + \Lambda x \right\rbrace - \dfrac{1}{2} \log |x| - \dfrac{1}{2}.
\end{equation}
Note that in the domain of definition of $\mathcal{R}_\mathbf{C}$, we have ${\mathcal{G}_\mathbf{C}'(x)=\frac{1}{2} \mathcal{R}_{\mathbf{C}}(x)}$. In this paper, $\mathcal{R}_ \mathbf{C}$ is applied to strictly negative values. For simplicity of notation, we will assume that it is well-defined in the considered range and use the R-transform notation. Otherwise, it can simply be replaced by $2 \mathcal{G}_\mathbf{C}'$, which is always valid.\\

The desired parameters $Q, q, m, \hat{Q}, \hat{q}, \hat{m}$ are solutions of the saddle-point equations, and describe the MAP estimator properties since we have maximized the integral over all distributions of $\mathbf{x}$. In particular, notice that the mean squared error $\tilde{E}$ and variance $\tilde{V}$ of the MAP estimator read at the saddle-point
\begin{align}
    \tilde{E} &= q - 2m + \mathbb{E}(x_0^2) \label{relation-E}\\
    \tilde{V} &= Q - q. \label{relation-V}
\end{align}
At this point, we differentiate the free energy \eqref{free_energy_RI} with respect to its parameters. For instance, differentiating with respect to $\hat{m}$ and using, \eqref{relation-V} gives
\begin{small}
\begin{equation}
    m = \int dx_0\phi(x_0) x_0 \int Dz \int \dfrac{dx}{\tilde{Z}}  x \exp \left\lbrace -\dfrac{f(x)}{\Delta} + \dfrac{x x_0}{\Delta} \mathcal{R}_{\mathbf{C}}\left(-\dfrac{\tilde{V}}{\Delta}\right) - \dfrac{x ^2}{2 \Delta}\mathcal{R}_{\mathbf{C}}\left(-\dfrac{\tilde{V}}{\Delta}\right) + z \sqrt{\hat{q}} x \right\rbrace \label{saddle-m}
\end{equation}
\end{small}
with $\tilde{Z} = \int dx \hspace{.1cm}   e^{-\frac{1}{\Delta}f(x)-\frac{\hat{Q}+\hat{q}}{2}x^2 + \hat{m} x x_0 + z \sqrt{\hat{q}}x}$. Combining saddle-point equations obtained by differentiating on $m$, $q$, and $Q-q$, and using \eqref{relation-E}, \eqref{relation-V} we also know
\begin{equation}
    \hat{q} = \dfrac{\Delta_0}{2 \Delta^2} \mathcal{R}_{\mathbf{C}}\left(-\dfrac{\tilde{V}}{\Delta}\right) + \dfrac{1}{2 \Delta} \left( \tilde{E} - \dfrac{\Delta_0}{\Delta}\tilde{V}\right)\mathcal{R}'_{\mathbf{C}}\left(-\dfrac{\tilde{V}}{\Delta}\right).
\end{equation}
which can be inserted into \eqref{saddle-m}. To take the limit $\Delta \rightarrow 0$, we rescale the variance parameter $\tilde{V}$ into $\tilde{V}/\Delta$, but keep the same name for simplicity. We are now left to do a Laplace approximation in the integral term of~\eqref{saddle-m}, to reach
\begin{footnotesize}
\begin{multline}
    m = \lim_{\Delta \rightarrow 0} \int dx_0 \phi(x_0) x_0 \int Dz \int \dfrac{dx}{\tilde{Z}} x \\
    \exp \left\lbrace - \dfrac{1}{\Delta} \left[ f(x) + \dfrac{x^2}{2} \mathcal{R}_{\mathbf{C}}(- \tilde{V})  - \dfrac{x x_0}{2} \mathcal{R}_{\mathbf{C}}(- \tilde{V}) - z x \sqrt{\Delta_0 \mathcal{R}_{\mathbf{C}}(- \tilde{V}) + (\tilde{E}-\Delta_0 \tilde{V})\mathcal{R}'_{\mathbf{C}}(- \tilde{V}) }\right] \right\rbrace \\
     =\mathbb{E}\left[ x_0 \argmin_{x}\left\lbrace f(x)+\dfrac{\mathcal{R}_{\mathbf{C}}(- \tilde{V})}{2} \left( x - \left[x_0 + 
     \dfrac{z}{\mathcal{R}_{\mathbf{C}} \left( -\tilde{V}\right)} \sqrt{\left( \tilde{E} - \Delta_0 \tilde{V} \right)\mathcal{R}_{\mathbf{C}}' \left( -\tilde{V}\right) + \Delta_0 \mathcal{R}_{\mathbf{C}} \left( -\tilde{V}\right)}\right]\right)^2\right\rbrace \right]. \label{prox-appear}
\end{multline}
\end{footnotesize}
Equation \eqref{prox-appear} clearly yields a proximal operator such that
\begin{footnotesize}
\begin{equation}
    m = \mathbb{E}\left[ x_0 \mbox{$\prox_{f/ \mathcal{R}_{\mathbf{C}}(-\tilde{V})}\left(x_{0} + \dfrac{z}{\mathcal{R}_{\mathbf{C}} \left( -\tilde{V}\right)} \sqrt{\left( \tilde{E} - \Delta_0 \tilde{V} \right)\mathcal{R}_{\mathbf{C}}' \left( -\tilde{V}\right) + \Delta_0 \mathcal{R}_{\mathbf{C}} \left( -\tilde{V}\right)}\right)$}\right].
\end{equation}
\end{footnotesize}
Similar computations on the saddle-point equations allow to rewrite them in terms of variables $(\tilde{E}, \tilde{V})$ and safely lead to \thref{theorem1}'s formulas:
\begin{footnotesize}
\begin{align}
    \tilde{V} &= \mathbb{E}\left[\dfrac{1}{\mathcal{R}_{\mathbf{C}}(-\tilde{V})}\mbox{$\prox'_{f/ \mathcal{R}_{\mathbf{C}}(-\tilde{V})}\left(x_{0} + \dfrac{z}{\mathcal{R}_{\mathbf{C}} \left( -\tilde{V}\right)} \sqrt{\left( \tilde{E} - \Delta_0 \tilde{V} \right)\mathcal{R}_{\mathbf{C}}' \left( -\tilde{V}\right) + \Delta_0 \mathcal{R}_{\mathbf{C}} \left( -\tilde{V}\right)}\right)$}\right] \label{SE_V}\\
    \tilde{E} &= \mathbb{E}\left[\left\lbrace \mbox{$\prox_{f/ \mathcal{R}_{\mathbf{C}}(-\tilde{V})}\left(x_{0} + \dfrac{z}{\mathcal{R}_{\mathbf{C}} \left( -\tilde{V}\right)} \sqrt{\left( \tilde{E} - \Delta_0 \tilde{V} \right)\mathcal{R}_{\mathbf{C}}' \left( -\tilde{V}\right) + \Delta_0 \mathcal{R}_{\mathbf{C}} \left( -\tilde{V}\right)}\right)-x_{0}$}\right\rbrace^{2}\right]. \label{SE_E}
\end{align}
\end{footnotesize}

Solving those equations, we obtain the mean squared error of the MAP estimator. However, the replica formula for this problem has not been rigorously justified yet. We conclude this section by connecting the fixed point of the state evolution equations of VAMP with the fixed point of the replica equations.

\begin{lemma}
\label{replica_SE_match}
The fixed point of the state evolution equations~\eqref{VAMP_se} yields the same MSE and variance as the replica prediction~\eqref{main-result}.
\end{lemma}

\emph{Proof.} See appendix \ref{appendix : rep_eq_se}.

\section{Properties of the proximal operator}
\subsection{Jacobian of the proximal}
\label{appendix : prox_prop}
The proximal operator can be written, for any parameter $\gamma \in \mathbb{R^{+}}$:
\begin{equation}
    \mbox{$\prox_{\gamma f}$}(\mathbf{x}) = \left(\rm{Id}+\gamma \partial f\right)^{-1}(\mathbf{x}).
\end{equation}
For any convex and differentiable function $f$, we have: 
\begin{equation}
    \mbox{$\prox_{\gamma f}$}(\mathbf{x})+\gamma \nabla f(\mbox{$\prox_{\gamma f}$}(\mathbf{x})) = \mathbf{x}
\end{equation}
For a twice differentiable $f$, applying the chain rule then yields :
\begin{equation}
    \mathcal{D}_{\mbox{$\prox_{\gamma f}$}}(\mathbf{x})+\gamma \mathcal{H}_{f}(\mbox{$\prox_{\gamma f}$}(\mathbf{x})) \mathcal{D}_{\mbox{$\prox_{\gamma f}$}}(\mathbf{x})  = Id
\end{equation}

where $\mathcal{D}$ is the Jacobian matrix and $\mathcal{H}$ the Hessian. Since f is a convex function, its Hessian is positive semi-definite, and, knowing that $\gamma$ is striclty positive, the matrix $(Id+\gamma \mathcal{H}_{f}(\mbox{$\prox_{\gamma f}$}(\mathbf{x})))$ is invertible. We thus have : 
\begin{equation}
    \mathcal{D}_{\mbox{$\prox_{\gamma f}$}}(\mathbf{x}) = (Id+\gamma \mathcal{H}_{f}(\mbox{$\prox_{\gamma f}$}(\mathbf{x})))^{-1}
\end{equation}

\subsection{Proximal of a sum}
\label{appendix : prox_prop2}
Although the identity $\mbox{$\prox_{f+g}$} = \mbox{$\prox_{f}$} \circ \mbox{$\prox_{g}$}$ does not always hold, it does when $g$ is a ridge penalty \cite{yu2013decomposing}. 
We then have, for any proper, convex, closed and separable function $f$:
\begin{equation}
    \mbox{$\prox_{f+\frac{\lambda_{2}}{2}\norm{.}_{2}^{2}}$} = \frac{1}{1+\lambda_{2}}\mbox{$\prox_{f}$}
\end{equation}
which allows to isolate the dependence of $\lambda_{2}$ in the replica equations:
\begin{scriptsize}
\begin{align*}
    \tilde{V} &= \mathbb{E}\left[\dfrac{1}{\mathcal{R}_{\mathbf{C}}(-\tilde{V})+\lambda_{2}}\mbox{$\prox'_{f/ \mathcal{R}_{\mathbf{C}}(-\tilde{V})}\left(x_{0} + \dfrac{z}{\mathcal{R}_{\mathbf{C}} \left( -\tilde{V}\right)} \sqrt{\left( \tilde{E} - \Delta_0 \tilde{V} \right)\mathcal{R}_{\mathbf{C}}' \left( -\tilde{V}\right) + \Delta_0 \mathcal{R}_{\mathbf{C}} \left( -\tilde{V}\right)}\right)$}\right] \\
    \tilde{E} &= \mathbb{E}\left[\left\lbrace \dfrac{1}{1+\frac{\lambda_{2}}{\mathcal{R}_{\mathbf{C}}(-\tilde{V})}}\mbox{$\prox_{f/\mathcal{R}_{\mathbf{C}}(-\tilde{V})}\left(x_{0} + \dfrac{z}{\mathcal{R}_{\mathbf{C}} \left( -\tilde{V}\right)} \sqrt{\left( \tilde{E} - \Delta_0 \tilde{V} \right)\mathcal{R}_{\mathbf{C}}' \left( -\tilde{V}\right) + \Delta_0 \mathcal{R}_{\mathbf{C}} \left( -\tilde{V}\right)}\right)-x_{0}$}\right\rbrace^{2}\right].
\end{align*}
\end{scriptsize}

\section{Main lemmas and \thref{theorem1}}

\subsection{Proof of Lemma~\ref{lemma3}}
\label{appendix : prox_fixed_point}
We start by reminding a useful identity on proximal operators:
\begin{proposition}(Resolvent of the sub-differential \cite{bauschke2011convex}) The proximal mapping of a convex function $f$ is the resolvent of the sub-differential of $f$:
    \begin{equation}
        \prox_{\gamma f} = (\rm{Id}+\gamma \partial f)^{-1}.
    \end{equation}
\end{proposition}
At the fixed point of the state evolution equations, we have $V_{1} = V_{2} = V$. Solving for the fixed point and replacing the proximal by the resolvent of $\partial f$ in \eqref{VAMP_alg}:
\begin{align}
    \mathbf{B}_{1} = A_{1} \hat{\mathbf{x}}_{1}+\partial f (\hat{\mathbf{x}}_{1}) \quad \quad \mathbf{B}_{2} = (\mathbf{F}^{T}\mathbf{F}+A_{2}\rm{Id})\hat{\mathbf{x}}_{2}-\mathbf{F}^{T}\mathbf{y}.
\end{align}
Replacing in the second line of \eqref{dr-vamp}  (either of the two equations):
\begin{equation}
    (\mathbf{F}^{T}\mathbf{F}+A_{2}\rm{Id})\hat{\mathbf{x}}_{2}-\mathbf{F}^{T}y = \frac{\hat{\mathbf{x}}_{1}}{V}-A_{1}\hat{\mathbf{x}}_{2}-\partial f (\hat{\mathbf{x}}_{1}).
\end{equation}
Knowing that $A_{1}+A_{2} = \frac{1}{V}$, and that $\hat{\mathbf{x}}_{1} = \hat{\mathbf{x}}_{2}$ at the VAMP fixed point, it reduces to: 
\begin{equation}
    \mathbf{F}^{T}(y-\mathbf{F}x) = \partial f ( \hat{\mathbf{x}}_{1})
\end{equation}
which is the optimality condition of problem \eqref{main-problem}. We see at the fixed point, the additional momentum terms and adaptative variances cancel out, giving the same fixed point as conventional proximal descent methods.

\subsection{Oracle-VAMP Lispchitz constants: $\ell_{2}$ case and proof of Lemma~\ref{lipschitz}}
\label{appendix : cvx_opt}
\subsubsection{A simple case: the $\ell_2$-penalty}
For the $\ell_2$-penalty case, oracle-VAMP's iteration~\eqref{oo} becomes a ridge regression with parameter $A_2$ which guarantees direct convergence of $\mathbf{B_2}$. Indeed, remember that the proximal operator of a $\ell_2$-penalty with parameter $\lambda_{2}$ is:
\begin{equation}
\label{prox_l2}
    \mbox{$\prox_{\frac{\lambda_{2}}{2}\vert \vert \mathbf{x} \vert \vert_{2}^{2}}$} = 1/(1+\lambda_{2}).
\end{equation}
Using \eqref{prox_l2} in \eqref{dr-vamp} immediately shows that $\mathbf{B}_{2}^{t}$ cancels itself, leading to the fixed point:
\begin{equation}
    \hat{\mathbf{x}}_{1} = \hat{\mathbf{x}}_{2} = (\mathbf{F}^{T}\mathbf{F}+A_{2}Id)^{-1}(\mathbf{F}^{T}\mathbf{y}).
\end{equation}

\subsubsection{General case}
We now turn to the general case and seek to establish Lipschitz bounds on operators $\mathcal{O}_1$ and $\mathcal{O}_2$. The approach is similar to that of \cite{giselsson2016linear} for Peaceman/Douglas-Rachford splitting. We start by reminding a few useful definitions from convex analysis.
\begin{definition}(Strong convexity) A proper closed function is $\sigma$-strongly convex with $\sigma >0$ if ${f-\frac{\sigma}{2}\norm{.}^{2}}$ is convex. If f is differentiable,
    the definition is equivalent to
    \begin{equation}
        f(x) \geqslant f(y) + \langle \nabla f(y), x-y \rangle + \frac{\sigma}{2} \norm{x-y}^{2}
    \end{equation}
    for all $x,y \in \mathcal{X}$.
\end{definition}

\begin{definition}(Smoothness for convex functions) A proper closed function $f$ is $\beta$-smooth with $\beta >0$ if $\frac{\beta}{2}\norm{.}^{2}-f$ is convex. If f is 
    differentiable, the definition is equivalent to 
    \begin{equation}
        f(x) \leqslant f(y) + \langle \nabla f(y), x-y \rangle + \frac{\beta}{2} \norm{x-y}^{2}
    \end{equation}
    for all $x,y \in \mathcal{X}$.
\end{definition}
An immediate consequence of those definitions is the following second order condition: for twice differentiable functions, $f$ is $\sigma$-strongly convex and $\beta$-smooth if and only if:
\begin{equation}
    \sigma \rm{Id} \preceq \mathcal{H}_{f} \preceq \beta \rm{Id}.
\end{equation}

\begin{corollary}(Remark 4.24 \cite{bauschke2011convex}) A mapping $T : \mathcal{X} \to \mathcal{D}$ (where $\mathcal{D}$ is a given output space) is $\beta$-cocoercive if and only if 
    $\beta$T is half-averaged. This means that T can be expressed as:
    \begin{equation}
        T = \frac{1}{2\beta}(\rm{Id}+S)
    \end{equation}
    where $S$ is a nonexpansive operator.
\end{corollary}

\par
The goal is now to determine the Lipschitz constants of $\mathcal{O}_{1}$ and $\mathcal{O}_{2}$ defined in \eqref{oo}.
\subsubsection{Lipschitz constant of $\mathcal{O}_{1}$}

\paragraph{Case 1: $0<\sigma_{1}<\beta_{1}$}
Proposition \ref{proposition : gisel_idh} gives the following expression:
\begin{align}
    \mbox{$\prox_{\frac{1}{A_{1}}f}$} &= \frac{1}{2}\left(\frac{1}{1+\sigma_{1}/A_{1}}+\frac{1}{1+\beta_{1}/A_{1}}\right)\rm{Id}+\frac{1}{2}\left(\frac{1}{1+\sigma_{1}/A_{1}}-\frac{1}{1+\beta_{1}/A_{1}}\right)S_{1}
\end{align}
where $S_{1}$ is a non-expansive operator. Replacing in the expression of $\mathcal{O}_{1}$ leads to:
\begin{align}
    \mathcal{O}_{1} &= \left(\frac{1}{2V}\left(\frac{1}{A_{1}+\sigma_{1}}+\frac{1}{A_{1}+\beta_{1}}\right)-1\right)\rm{Id}+\frac{1}{2V}\left(\frac{1}{1+\sigma_{1}/A_{1}}-\frac{1}{1+\beta_{1}/A_{1}}\right)S_{1}\left(\frac{.}{A_{1}}\right)
\end{align}
which, knowing that $A_{1}+A_{2} = \frac{1}{V}$, $\mathcal{O}_{1}$ has Lipschitz constant:
\begin{equation}
    \mathcal{L}_{1} = \max \left(\frac{A_{2}-\sigma_{1}}{A_{1}+\sigma_{1}},\frac{\beta_{1}-A_{2}}{A_{1}+\beta_{1}}\right).
\end{equation}

\paragraph{Case 2: $0<\sigma_{1}=\beta_{1}$}

In this case, we have from Proposition \ref{proposition : gisel_idh}:
\begin{equation}
    \norm{\mbox{$\prox_{\frac{1}{A_{1}}f}$}(x)-\mbox{$\prox_{\frac{1}{A_{1}}f}$}(y)}_{2}^{2} = \left(\frac{1}{1+\sigma_{1}/A_{1}}\right)^{2} \norm{x-y}_{2}^{2}
\end{equation}
which, with the firm non-expansiveness of the proximal operator gives:
\begin{align}
    \norm{\mathcal{O}_{1}(x)-\mathcal{O}_{1}(y)}_{2}^{2} &= \frac{1}{V^{2}} \norm{\mbox{$\prox_{\frac{1}{A_{1}}f}$}(x/A_{1})-\mbox{$\prox_{\frac{1}{A_{1}}f}$}(y/A_{1})}_{2}^{2}\\
    &-2\frac{A_{1}}{V}\left\langle \frac{x}{A_{1}}-\frac{y}{A_{1}},\mbox{$\prox_{\frac{1}{A_{1}}f}$}(x/A_{1})-\mbox{$\prox_{\frac{1}{A_{1}}f}$}(y/A_{1})\right\rangle+\norm{x-y}_{2}^{2} \\
    & \label{equation : stuff} \leqslant \left(\frac{1}{V^{2}}-2\frac{A_{1}}{V}\right)\norm{\mbox{$\prox_{\frac{1}{A_{1}}f}$}(x/A_{1})-\mbox{$\prox_{\frac{1}{A_{1}}f}$}(y/A_{1})}_{2}^{2} +\norm{x-y}_{2}^{2} \\
    &= \left(\left(\frac{1}{V^{2}}-2\frac{A_{1}}{V}\right)\left(\frac{1}{A_{1}+\sigma_{1}}\right)^{2}+1\right)\norm{x-y}_{2}^{2} \\
    &= \left(\frac{A_{2}^{2}-A_{1}^{2}}{(A_{1}+\sigma_{1})^{2}}+1\right)\norm{x-y}_{2}^{2}.
\end{align}
The upper bound on the Lipschitz constant is therefore:
\begin{equation}
    \mathcal{L}_{1} = \sqrt{\frac{(A_{2}^{2}-A_{1}^{2})}{(A_{1}+\sigma_{1})^{2}}+1}.
\end{equation}

\paragraph{Case 3: no strong convexity or smoothness assumption}

In this case the only information we have is the firm nonexpansiveness of the proximal operator, which gives the same proof as in the previous case but stops at \eqref{equation : stuff}, immediately giving the upper bound:
\begin{equation}
    \mathcal{L}_{1} = \max \left(1,\frac{{A}_{1}}{A_{2}}\right).
\end{equation}

\paragraph{Lipschitz constant of $\mathcal{O}_{2}$}

Remember that we make the assumption that the data matrix is non-trivial, i.e. that $\lambda_{max}(\mathbf{F}^T \mathbf{F}) \neq 0$. In this case we use the explicit form of $\mathcal{O}_{2}$, which is linear: 
\begin{align}
    \label{equation : spectral_norm}
    \norm{\mathcal{O}_{2}(x)-\mathcal{O}_{2}(y)}_{2} &= \norm{\left( \frac{1}{V}(\mathbf{F}^{T}\mathbf{F}+A_{2}^{t}\rm{Id})^{-1}-\mathbf{I}\right)(x-y)}_{2} \\
    & \leqslant \norm{\left( \frac{1}{V}(\mathbf{F}^{T}\mathbf{F}+A_{2}^{t}\rm{Id})^{-1}-\mathbf{I}\right)}_{2} \hspace{.1cm} \norm{x-y}_{2}.
\end{align}
The spectral norm of the matrix in \eqref{equation : spectral_norm} gives the upper bound on the Lipschitz constant:
\begin{equation}
    \mathcal{L}_{2} = \max \left(\frac{A_{1}-\lambda_{min}(\mathbf{F}^{T}\mathbf{F})}{A_{2}+\lambda_{min}(\mathbf{F}^{T}\mathbf{F})}, \frac{\lambda_{max}(\mathbf{F}^{T}\mathbf{F})-A_{1}}{A_{2}+\lambda_{max}(\mathbf{F}^{T}\mathbf{F})} \right)
\end{equation}
\subsection{Proof of Lemma \ref{stieltjes-lemma}}
\label{appendix : stieltjes-lemma-proof}
The equation defining $V_{2}$ in \eqref{VAMP_alg} directly gives $V = \mathcal{S}_{\mathbf{C}}(-A_{2})$ by the definition of the Stieltjes transform. For a separable and differentiable function, we have the element-wise identity (see appendix \ref{appendix : prox_prop})
\begin{equation}
     \mbox{$\prox^{'}_{\gamma f}$}(x) = \frac{1}{1+\gamma f^{''}( \mbox{$\prox_{\gamma f}$}(x))}
\end{equation}
which, from the definition of the element wise averaging operator, gives:
\begin{equation}
    \langle \mbox{$\prox^{'}_{\gamma f}$}(x) \rangle = \frac{1}{N}\mbox{Trace}\left[(\rm{Id}+\gamma\mathcal{H}_{f}(\hat{\mathbf{x}}))^{-1}\right].
\end{equation}
The prescription for $V_{1}$ in \eqref{VAMP_alg} then directly gives $V = \mathcal{S}_{\mathcal{H}_{f}(\hat{\mathbf{x}})}(-A_{1})$.

\subsection{Proof of Lemma \ref{lemma-bounds}}
\label{appendix : proof-lemma-bounds}
From the definition of the Stieltjes transform and the second order definition of strong convexity and smoothness:
\begin{equation}
    \frac{1}{\lambda_{max}(\mathbf{F}^{T}\mathbf{F})+A_{2}} \leqslant \frac{1}{V} \leqslant \frac{1}{\lambda_{min}(\mathbf{F}^{T}\mathbf{F})+A_{2}} \quad \quad \frac{1}{\sigma_{1}+A_{1}} \leqslant \frac{1}{V} \leqslant \frac{1}{\beta_{1}+A_{1}} 
\end{equation}
which combined with $A_{1}+A_{2} = \frac{1}{V}$ gives the bound. 

\subsection{Proof of Lemma~\ref{replica_SE_match} and \thref{theorem1}}
\label{appendix : rep_eq_se}
The state evolution equations~\eqref{VAMP_se} follow a set of parameters through the VAMP iterations. Among those parameters, we find the mean squared errors $\mathcal{E}_1$ and $\mathcal{E}_2$ of the estimators $\mathbf{\hat{x}_1}$ and $\mathbf{\hat{x}_2}$ with respect to the true signal, and their variances $V_1$, $V_2$. At the fixed-point of VAMP, the estimators are equal, hence their errors and variances coincide with one mean squared error $E$ and variance $V$. Besides, the replica saddle point equations also close on the error $\tilde{E}$ and variance $\tilde{V}$. We would like to show that the replica prediction matches the state evolution fixed point conditions.

Starting from the state evolution fixed point, we notice that
\begin{equation}
    V = \dfrac{\alpha_2}{A_2} = \mathcal{S}_{\mathbf{C}}(-A_2)
\end{equation}
where $S$ is the Stieltjes transform with respect to the spectral measure defined by $ \mathbf{C} = \mathbf{F}^T \mathbf{F}$.
Then
\begin{equation}
    A_1 = \dfrac{1}{V} - A_2 = \dfrac{1}{V} + \mathcal{S}_{\mathbf{C}}^{-1}(-V) = \mathcal{R}_{\mathbf{C}}(-V).
\end{equation}
Moreover, 
\begin{equation}
    V = \dfrac{1}{A_1} \mathbb{E}_{x_0, P_1} \left[ \mbox{$\prox$}'_{f/A_1}(x_0 + P_1) \right] \label{V-se}
\end{equation}
where $P_1$ is a Gaussian variable of variance $\tau_1$.
Looking at $\tau_2$, we have
\begin{equation}
    \tau_2=\dfrac{1}{(1-A_1 V)^2}\left[E-\tau_1 (A_1 V)^2\right]=\dfrac{1}{(A_2 V)^2}\left[E-\tau_1 (1- A_2 V)^2\right].
    \label{tau2_VAMP}
\end{equation}
We rewrite the equation on $\tau_1$, which involves an average on the eigenvalue distribution of ${\mathbf{C} = \mathbf{F}^T \mathbf{F}}$:
\begin{align}
    \tau_1 &= \frac{1}{(1-A_2 V)^2}
\left({\mathbb E}\left[\Delta_0 \frac{\lambda}{(\lambda + A_2 )^2}  + \tau_2 \frac{ A_2^2}{(\lambda + A_2 )^2} \right]  - \tau_2 (A_2 V)^2\right) \\
 &= \frac{1}{(1-A_2 V)^2}
\left({\mathbb E}\left[\frac{\Delta_0}{(\lambda + A_2)} -\frac{A_2 \Delta_0 }{(\lambda + A_2 )^2}  + \tau_2 \frac{ A_2^2}{(\lambda + A_2 )^2} \right]  - \tau_2 (A_2 V)^2\right) \\
\tau_1&= \frac{1}{(1-A_2 V)^2} \left(\Delta_0 \mathcal{S}_{\mathbf{C}}(-A_2) - A_2 \Delta_0  \mathcal{S}_{\mathbf{C}}'(-A_2 ) + \tau_2  A_2^2\mathcal{S'}_{\mathbf{C}}(-A_2 ) - \tau_2 (A_2 V)^2\right),
\end{align}
then plug in $\tau_2$'s expression~\eqref{tau2_VAMP} to reach
\begin{align}
    \tau_1 &= \dfrac{\Delta_0 V^2}{ (1-A_2 V)^2 \mathcal{S}_{\mathbf{C}}'(- A_2)} (\mathcal{S}_{\mathbf{C}}(- A_2) -  A_2 \mathcal{S}_{\mathbf{C}}'(- A_2)) \nonumber \\
    &+ \dfrac{E}{ (1-A_2 V)^2 \mathcal{S}_{\mathbf{C}}'(- A_2)}( \mathcal{S}_{\mathbf{C}}'(- A_2) - V^2). \label{tau1_VAMP}
\end{align}
We have expressed the variance of the Gaussian variable $P_1$ as a function of $E$ and $V$.
We would like to match this with the variance of the Gaussian inside~\eqref{SE_V}, namely
\begin{align}
\tilde{\tau}_1 &= \frac{1}{\mathcal{R}_{\mathbf{C}}^2\left(-\frac{\tilde{V}}{\Delta}\right)} \left(\left(\tilde{E}-\dfrac{\Delta_0}{\Delta}\tilde{V}\right) \mathcal{R}_{\mathbf{C}}'\left(-\frac{\tilde{V}}{\Delta}\right)  + \Delta_0 \mathcal{R}_{\mathbf{C}}\left(-\frac{\tilde{V}}{\Delta}\right)\right) \\
 &= \dfrac{E}{\mathcal{R}_{\mathbf{C}}^2\left(-\frac{\tilde{V}}{\Delta}\right)}\left(\frac {-1}{\mathcal{S}_{\mathbf{C}}'(\mathcal{S}_{\mathbf{C}}^{-1}(\frac{\tilde{V}}{\Delta}))} + \frac {\Delta^2}{\tilde{V}^2}\right) \nonumber \\
 &+ \dfrac{\Delta_0}{\mathcal{R}_{\mathbf{C}^2}\left(-\frac{\tilde{V}}{\Delta}\right)} \left( \mathcal{R}_{\mathbf{C}}\left(-\frac{\tilde{V}}{\Delta}\right) - \dfrac{\tilde{V}}{\Delta}\left(\frac {-1}{\mathcal{S}_{\mathbf{C}}'(\mathcal{S}_{\mathbf{C}}^{-1}(\frac{\tilde{V}}{\Delta}))} + \frac {\Delta^2}{\tilde{V}^2}\right) \right). \label{tau1_replica}
 \end{align}
A few lines of computation show that $\tau_1 = \tilde{\tau}_1$. Therefore, the replica saddle point equation on $\tilde{V}$ \eqref{V-se} becomes exactly the same as the state evolution fixed point equation on $V$ \eqref{V-replica}. Similarly, we recall the definition of the fixed point value of E for SE equations
\begin{equation}
    E = \mathbb{E}\left[\left( \mbox{$\prox$}_{\frac{1}{A_{1}}f}(x_{0}+P_{1})-x_{0} \right)^{2}\right] \label{E-se}
\end{equation}
where $P_1$ is also pulled from a Gaussian distribution with variance $\tau_1 = \tilde{\tau}_1$. \eqref{E-se} matches the replica equation on $\tilde{E}$ \eqref{E-replica}. Finally the variables $(E, V)$ from SE equations, and $(\tilde{E}, \tilde{V})$ from replica formalism are the same and satisfy the same relations, which proves Lemma~\ref{replica_SE_match}. \\It is hence straightforward to prove \thref{theorem1} after having shown \thref{theorem2}. As shown above, \thref{theorem1} is simply a rewriting of state evolution equations from \thref{theorem2} in their replica fixed point form, i.e. removing some intermediate variables to obtain a more compact form. 

\section{State evolution equations for the elastic net problem}
\label{appendix : e_net}
We solve the recursion \eqref{VAMP_alg} on an elastic net problem:
\begin{equation}
    \hat{\mathbf{x}} = \argmin_{\mathbf{x}\in\mathbb{R}^{N}}\left\lbrace \frac{1}{2} \norm{\mathbf{y}-\mathbf{F}\mathbf{x}}_{2}^{2}+\lambda_{1} \vert \mathbf{x} \vert_{1}+\frac{\lambda_{2}}{2}\norm{\mathbf{x}}_{2}^{2} \right\rbrace.
\end{equation}
For a given parameter $\gamma \in \mathbb{R}^{+}$, the proximal operator of the corresponding regularization function reads:
\begin{equation}
    \mbox{$\prox_{\frac{1}{A_{1k}}(\lambda_{1} \vert \mathbf{x} \vert_{1}+\frac{\lambda_{2}}{2}\norm{\mathbf{x}}_{2}^{2})}$}(.) = \frac{1}{1+\frac{\lambda_{2}}{A_{1k}}}\hspace{.1cm}s\hspace{-0.1cm}\left(.,\frac{\lambda_{1}}{A_{1k}}\right)
\end{equation}
where $s\left(.,\frac{\lambda_{1}}{A_{1k}}\right)$ is the soft-thresholding function: 
\begin{equation}
s\left(r_{1k},\frac{\lambda_{1}}{A_{1k}}\right) =
\left\lbrace
\begin{array}{ccc}
r_{1k}+\frac{\lambda_{1}}{A_{1k}}  & \mbox{if} & r_{1k}<-\frac{\lambda_{1}}{A_{1k}}\\
0 & \mbox{if} & -\frac{\lambda_{1}}{A_{1k}}<r_{1k}<\frac{\lambda_{1}}{A_{1k}}\\
r_{1k}-\frac{\lambda_{1}}{A_{1k}} & \mbox{if} & r_{1k}>\frac{\lambda_{1}}{A_{1k}}.
\end{array}\right.
\end{equation}
We consider an i.i.d. teacher vector $\mathbf{x}_{0}$ pulled from the Gauss-Bernoulli distribution :
\begin{equation}
    \phi(x_{0}) = (1-\rho)\delta(x_{0})+\rho\frac{1}{\sqrt{2\pi}}\exp{(-x_{0}^{2}/2)}.
\end{equation}
Here we give the detail of the set of equations \eqref{VAMP_se} for an elastic net minimization problem. The quantities that must be explicitly computed are the averages $\mathcal{E}_{1}$ and $\mathcal{E}_{2}$ and the ones on the derivatives.

\begin{align}
    \alpha_{1k} &= \mathbb{E}\left[\frac{1}{1+\frac{\lambda_{2}}{A_{1k}}}s'\left(x_{0}+P_{1k},\frac{\lambda_{1}}{A_{1k}}\right)\right] \quad \mbox{where} \quad x_{0} \sim \mathcal{N}(0,1) \quad p_{1k} \sim \mathcal{N}(0,\tau_{1k}) \\
    &=\frac{1}{1+\frac{\lambda_{2}}{A_{1k}}}(1-\rho)\left(\int_{-\infty}^{-\lambda_{1}/A_{1k}}dp\frac{1}{\sqrt{2\pi \tau_{1k}}}e^{-\frac{p^2}{2\tau_{1k}}}+\int_{\lambda_{1}/A_{1k}}^{+\infty}dp\frac{1}{\sqrt{2\pi \tau_{1k}}}e^{-\frac{p^{2}}{2\tau_{1k}}}\right) \\
    &+\rho\frac{1}{1+\frac{\lambda_{2}}{A_{1}}}\bigg(\int_{-\infty}^{-\lambda_{1}/A_{1k}}dp\frac{1}{\sqrt{2\pi(\tau_{1k}+1)}}\exp(-\frac{p^2}{2(\tau_{1k}+1)}) \notag \\
    &+\int_{\lambda_{1}/A_{1k}}^{+\infty}dp\frac{1}{\sqrt{2\pi(\tau_{1k}+1)}}\exp\left(-\frac{p^{2}}{2(\tau_{1k}+1)}\right)\bigg) \notag \\
    &= \frac{1}{1+\frac{\lambda_{2}}{A_{1k}}}\left[(1-\rho)\erfc\left(\frac{\lambda_{1}}{A_{1k}\sqrt{2\tau_{1k}}}\right)+\rho\erfc\left(\frac{\lambda_{1}}{A_{1k}\sqrt{2(\tau_{1k}+1)}}\right)\right].
\end{align}

\begin{align}
    \mathcal{E}_{1} &= \mathbb{E}_{x_{0},P_{1k}}\left[\left(\mbox{$\prox_{\frac{1}{A_{1k}}f}(x_{0}+P_{1k})-x_{0}$}\right)^2\right]\quad \mbox{where } x_{0} \sim \mathcal{N}(0,1) \quad p_{1k} \sim \mathcal{N}(0,\tau_{1k})\\
    &=\left(\frac{1}{1+\frac{\lambda_{2}}{A_{1k}}}\right)^{2}(1-\rho)\int_{\mathbb{R}}dp\frac{1}{\sqrt{2\pi \tau_{1k}}}\exp\left(-\frac{p^{2}}{2\tau_{1k}}\right)s\left(p,\frac{\lambda_{1}}{A_{1k}}\right)^{2} \label{equation : p1_e1}\\
    \label{equation : p2_e1}+\rho&\iint_{\mathbb{R}}dx_{0}dp\frac{1}{\sqrt{2\pi \tau_{1k}}}\exp\left(-\frac{p^{2}}{2\tau_{1k}}\right)\frac{1}{\sqrt{2\pi}}\exp\left(-\frac{x_{0}^{2}}{2}\right)\left(\frac{1}{1+\frac{\lambda_{2}}{A_{1k}}}s\left(x_{0}+p,\frac{\lambda_{1}}{A_{1k}}\right)-x_{0}\right)^{2}. 
\end{align}
The first term \eqref{equation : p1_e1} only involves one dimensional integrals and can easily be evaluated numerically:
\begin{small}
\begin{multline}
    \hspace{-0.5cm}\eqref{equation : p1_e1} = (1-\rho)\left(\frac{1}{1+\frac{\lambda_{2}}{A_{1k}}}\right)^{2}\left[\int_{-\infty}^{-\lambda_{1}/A_{1k}}dp\mathcal{N}(0,\tau_{1k})\left(p+\frac{\lambda_{1}}{A_{1k}}\right)^{2}+\int_{\lambda_{1}/A_{1k}}^{+\infty}dp\mathcal{N}(0,\tau_{1k})\left(p-\frac{\lambda_{1}}{A_{1k}}\right)^{2}\right] \\
    =(1-\rho)\left(\frac{1}{1+\frac{\lambda_{2}}{A_{1k}}}\right)^{2}\left[\erfc{\left(\frac{\lambda_{1}/A_{1k}}{\sqrt{2\tau_{1k}}}\right)}\left(\left(\frac{\lambda_{1}}{A_{1k}}\right)^{2}+\tau_{1k}\right)-e^{-\frac{(\lambda_{1}/A_{1k})^{2}}{2\tau_{1k}}}\sqrt{2\tau_{1k}/\pi} \frac{\lambda_{1}}{A_{1k}}\right].
\end{multline}
\end{small}
The second term \eqref{equation : p2_e1} needs to be decomposed in order to avoid computing the two-dimensional integral numerically:
\begin{multline}
    \eqref{equation : p2_e1} = \rho\mathbb{E}_{x_{0}}\bigg[\int_{-\infty}^{-\lambda_{1}/A_{1k}-x_{0}}dp\mathcal{N}(0,\tau_{1k})\left(\frac{1}{1+\frac{\lambda_{2}}{A_{1k}}}\left(x_{0}+p+\frac{\lambda_{1}}{A_{1k}}\right)-x_{0}\right)^{2} \\
    +\int_{\lambda_{1}/A_{1k}-x_{0}}^{+\infty}dp\mathcal{N}(0,\tau_{1k})\left(\frac{1}{1+\frac{\lambda_{2}}{A_{1k}}}\left(x_{0}+p-\frac{\lambda_{1}}{A_{1k}}\right)-x_{0}\right)^{2} +\int_{-\lambda_{1}/A_{1k}-x_{0}}^{\lambda_{1}/A_{1k}-x_{0}}dp\mathcal{N}(0,\tau_{1k})x_{0}^{2}\bigg]. \notag
\end{multline}
\\
A little algebra allows to express \eqref{equation : p2_e1} with error functions supported by most scientific coding libraries. We rewrite the shrinkage factor due to the $\ell_2$ penalty $s = \frac{1}{1+\frac{\lambda_{2}}{A_{1k}}}$:

\begin{align}
    \eqref{equation : p2_e1} &= \rho\mathbb{E}_{x_{0}}\bigg[\frac{1}{2}x_{0}^2\left(\erf\left(\frac{\lambda_{1}/A_{1k}-x_{0}}{\sqrt{2\tau_{1k}}}\right)+\erf\left(\frac{\lambda_{1}/A_{1k}+x_{0}}{\sqrt{2\tau_{1k}}}\right)\right) \\
    &+x_{0}^{2}-2sx_{0}^{2}+s^{2}(\tau_{1k}+(\lambda_{1}/A_{1k})^{2}+x_{0}^{2}) \\
    &+s\sqrt{\tau_{1k}/(2\pi)}\bigg(\exp{\left(-\frac{\lambda_{1}/A_{1k}-x_{0}^{2}}{2\tau_{1k}}\left((s-2)x_{0}-s\frac{\lambda_{1}}{A_{1k}}\right)\right)}\\
    & +\exp{\left(-\frac{\lambda_{1}/A_{1k}+x_{0}^{2}}{2\tau_{1k}}\left((2-s)x_{0}-s\frac{\lambda_{1}}{A_{1k}}\right)\right)}\bigg) \\
    &+\frac{1}{2}\bigg((s^{2}(\tau_{1k}+(\lambda_{1}/A_{1k}-x_{0})^{2})+2s(\lambda_{1}/A_{1k}-x_{0})x_{0}+x_{0}^{2})\erf{\left(\frac{\lambda_{1}/A_{1k}-x_{0}}{\sqrt{2\tau_{1k}}}\right)} \\
    &-(x_{0}^{2}-2sx_{0}(\lambda_{1}/A_{1k}+x_{0})+s^{2}(\tau_{1k}+(\lambda_{1}/A_{1k}+x_{0})^{2}))\erf{\left(\frac{\lambda_{1}/A_{1k}+x_{0}}{\sqrt{2\tau_{1k}}}\right)}\bigg)\bigg].
\end{align}

We then invoke the appropriate expressions for $\alpha_{2k}$ and $\mathcal{E}_{2k}$ from \cite{rangan2019vector}:
\begin{align}
    \alpha_{2k} &= \mathbb{E}\left[\frac{A_{2k}}{\lambda_{\mathbf{F}^{T}\mathbf{F}}+A_{2k}}\right] \\
    \mathcal{E}_{2} &= \mathbb{E}\left[\frac{\Delta_{0}\lambda_{\mathbf{F}^{T}\mathbf{F}}+\tau_{2k} A_{2k}^{2}}{(\lambda_{\mathbf{F}^{T}\mathbf{F}}+A_{2k})^{2}}\right].
\end{align}
These forms can be used in the recursion \eqref{VAMP_se} with the chosen values of $\lambda_{1}$ and $\lambda_{2}$ to find the right thresholding coefficients, errors and variances. \\

\begin{figure}
\begin{center}
\includegraphics[scale=0.55]{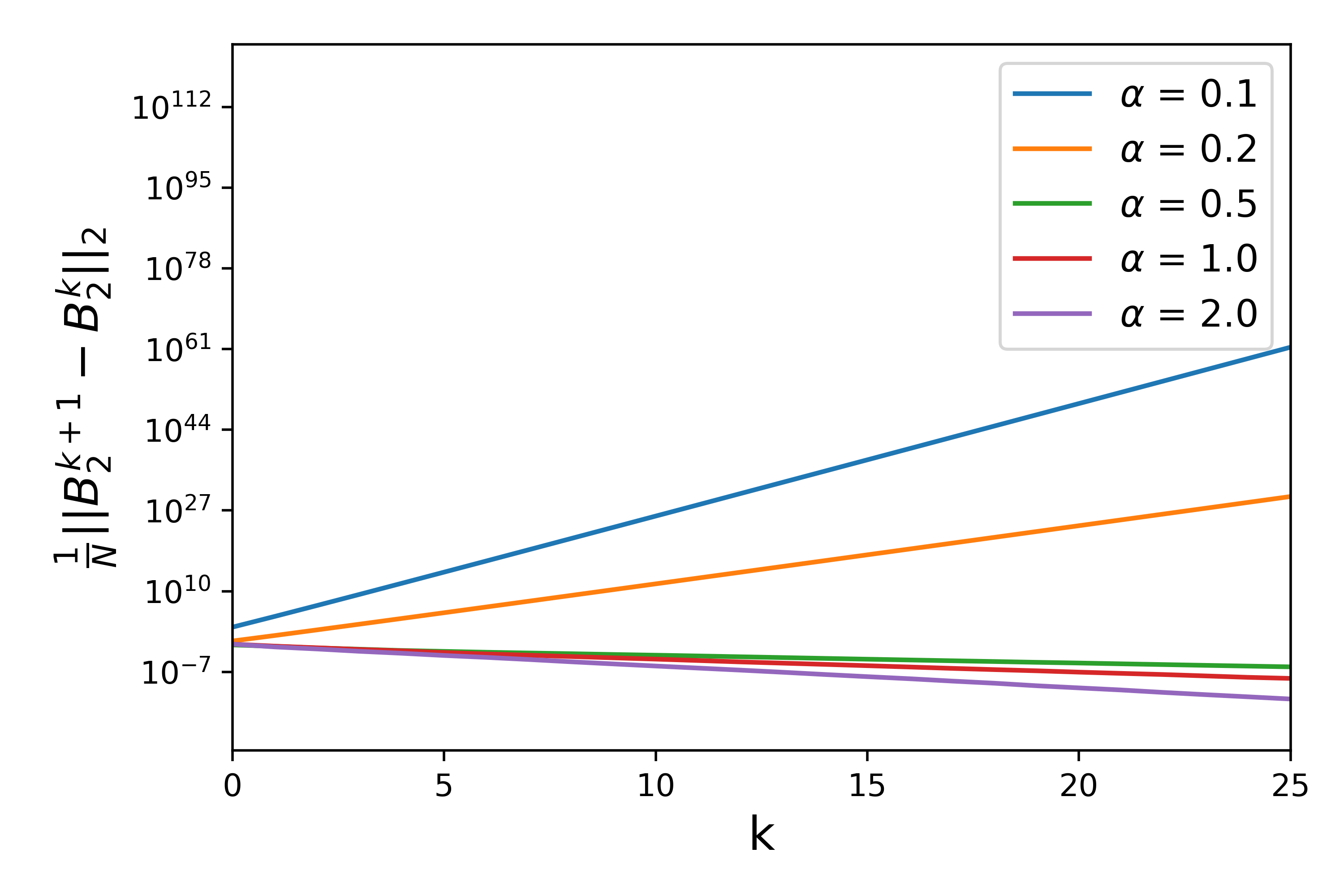}
\includegraphics[scale=0.55]{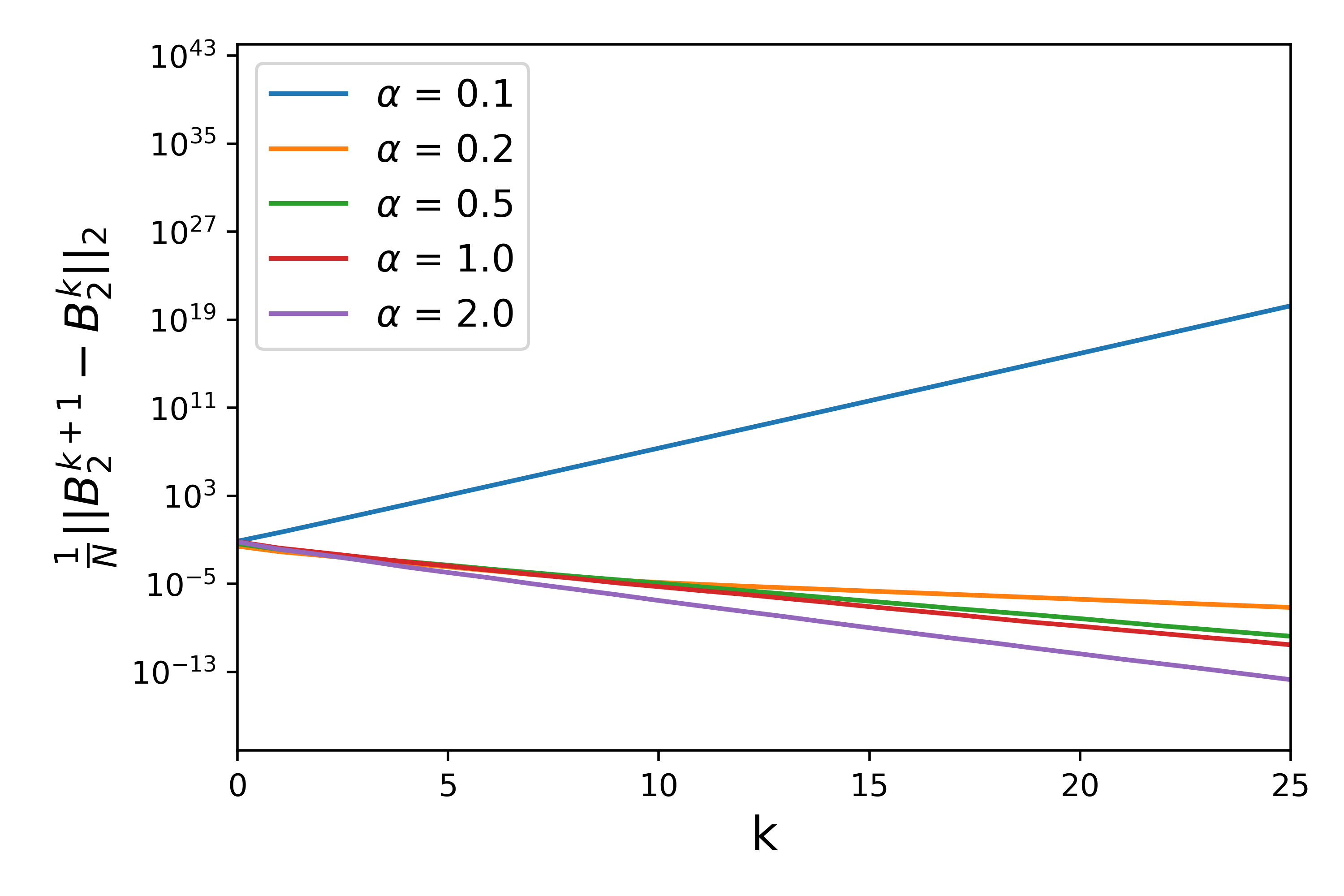}
\includegraphics[scale=0.55]{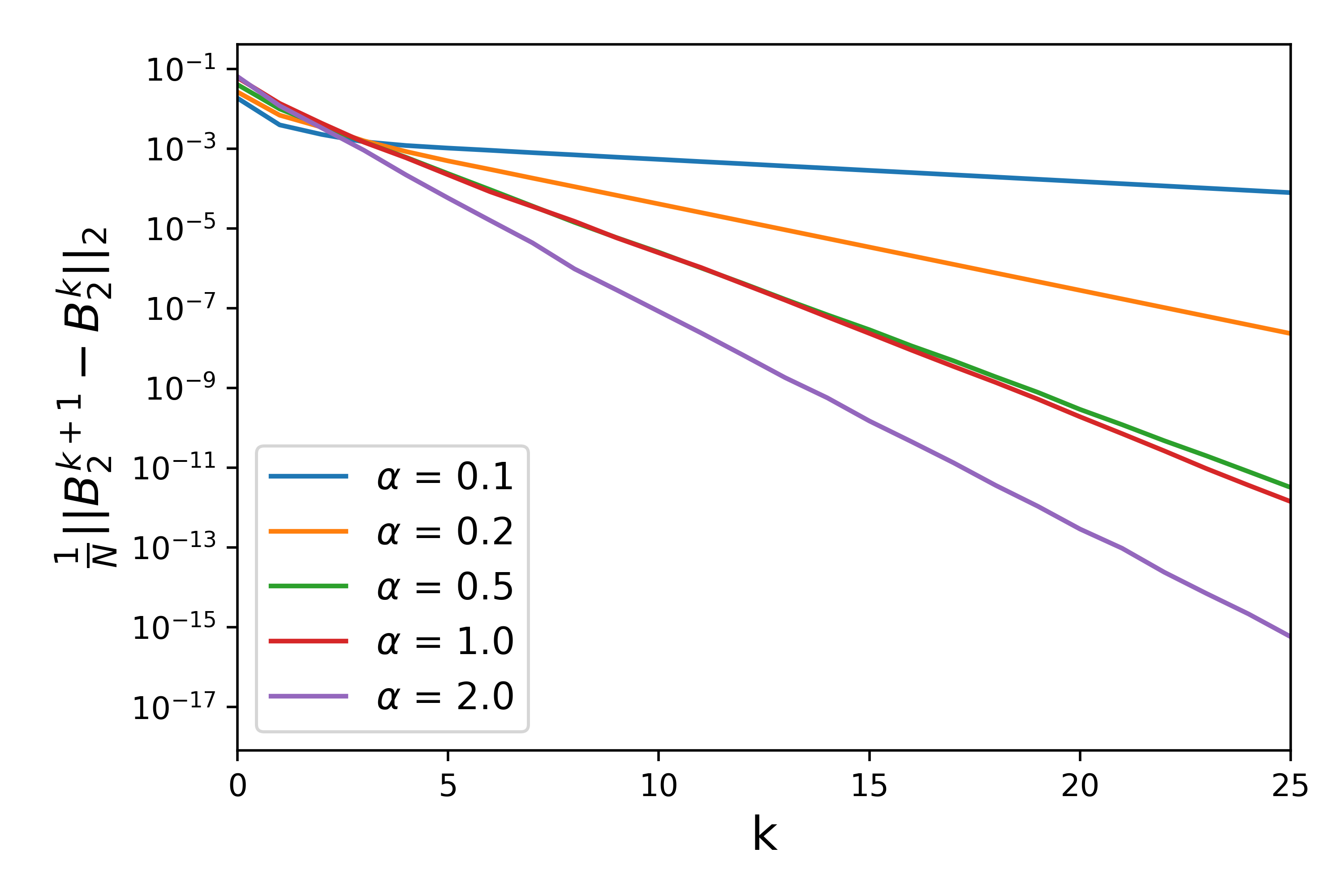}
\caption{Logarithmic scale mean squared distance between successive iterates of oracle-VAMP on an elastic net problem with $\lambda_{1} = 0.1$ and, from left to right $\lambda_{2} = 0.1,0.2,0.3$. Each plot contains five trajectories with aspect ratio $\alpha = 0.1,0.2,0.5,1.0,2.0$. For low aspect ratios, the sensing matrix is highly ill-conditioned and diverging trajectories are observed (blue $\alpha=0.1$ and orange $\alpha=0.2$ curves, on the first plot). Augmenting the ridge parameter enforces the convergence: at $\lambda_{2} = 0.2$ (second plot), the orange curve describes a converging trajectory, and at $\lambda_{2} = 0.3$ (third plot), the blue curve converges as well. Plots are obtained with $N = 100$, $\rho = 0.3$ and $\Delta_{0} = 0.01$. The plots were generated using the toolbox from \url{https://github.com/cgerbelo/Oracle_VAMP.}}
\label{fig:nice_plot_from_heaven}
\end{center}
\end{figure}

We used these forms with $\lambda_{2} = 0$ for the LASSO simulations in the experiments in section \ref{subsection : experiments}. The elastic net also allows us to illustrate the convergence for large enough $\lambda_{2}$, as shown in figure \ref{fig:nice_plot_from_heaven}. We launch oracle-VAMP on the elastic net problem for five values of the aspect ratio ${\alpha = 0.1,0.2,0.5,1,2}$. The problem setup is the same as in \ref{subsection : experiments}, with $\rho = 0.3$, $\Delta_{0} = 0.01$. The choice of sensing matrix matters little here, as long as the eigenvalue spectrum has compact support. We used i.i.d. Gaussian matrices for simplicity. We plot the average squared distance between two successive iterates of \eqref{dr-vamp} in a logarithmic scale on an elastic net problem with a LASSO parameter of $\lambda_{1} = 0.1$ and varying ridge parameter $\lambda_{2} = 0.1,0.2,0.3$. For low values of $\alpha$, the data matrix is highly ill-conditioned and the algorithm diverges as shown on the first plot for $\alpha = 0.1, 0.2$. We then augment the ridge parameter on the second figure, which makes the $\alpha = 0.2$ curve converge. Pushing $\lambda_{2}$ further on the third plot makes the $\alpha = 0.1$ curve converge. We thus see that augmenting the ridge indeed enforces convergence. 

\section{Convergence of vector sequences}
\label{appendix:analysis_framework}
This is essentially a rewriting of appendix B of \cite{rangan2019vector}, which reviews the analysis framework from \cite{bayati2011dynamics}. \\ The main building blocks are the notions of \emph{vector sequence} and \emph{pseudo-Lipschitz function}, which allow to define the \emph{empirical convergence with p-th order moment}.
Consider a vector of the form
\begin{equation}
    \mathbf{x}(N) = (\mathbf{x}_{1}(N),...,\mathbf{x}_{N}(N))
\end{equation}
where each sub-vector $\mathbf{x}_{n}(N) \in \mathbb{R}^{r}$ for any given $r \in \mathbb{N}^{*}$. For r=1, which we use in \thref{theorem1}, $\mathbf{x}(N)$ is denoted a \emph{vector sequence}. \\
Given $p\geqslant 1$, a function $\mathbf{f} :\mathbb{R}^{r}\to \mathbb{R}^{s}$ is said to be \emph{pseudo-Lipschitz continuous of order p} if there exists a constant $C>0$ such that for all $\mathbf{x}_{1}, \mathbf{x}_{2} \in \mathbb{R}^{s}$:
\begin{equation}
    \norm{\mathbf{f}(\mathbf{x}_{1})-\mathbf{f}(\mathbf{x}_{2})} \leqslant C \norm{\mathbf{x}_{1}-\mathbf{x}_{2}}\left[1+\norm{\mathbf{x}_{1}}^{p-1}+\norm{\mathbf{x}_{1}}^{p-1}\right]
\end{equation}
Then, a given vector sequence $\mathbf{x}(N)$ \emph{converges empirically with p-th order moment} if there exists a random variable $X \in \mathbb{R}^{r}$ such that:
\begin{itemize}
    \item $\mathbb{E}\abs{X}^{p}<\infty$; and
    \item for any scalar-valued pseudo-Lipschitz continuous $\mathbf{f}(.)$ of order p,
    \begin{equation}
        \lim_{N\to \infty}\frac{1}{N}\sum_{n=1}^{N}\mathbf{f}(x_{n}(N))=\mathbb{E}[f(X)] \thickspace \mbox{a.s.}
    \end{equation}
\end{itemize}
Note that defining an empirically converging singular value distribution implicitly defines a sequence of matrices $\mathbf{F}(N)$ using the definition of rotational invariance from the introduction. This naturally brings us back to the original definitions from \cite{bayati2011dynamics}.
An important point is that the almost sure convergence of the second condition holds for random vector sequences, such as the ones we consider in the introduction. We also remind the definition of \emph{uniform Lipschitz continuity}.\\

For a given mapping $\phi(\mathbf{x},A)$ defined on $\mathbf{x} \in \mathcal{X}$ and $A \in \mathbb{R}$, we say it is \emph{uniform Lipschitz continuous} in $\mathbf{x}$ at $A = \bar{A}$ if there exists constants $L_{1}$ and $L_{2} \geqslant 0$ and an open neighborhood U of $\bar{A}$ such that:
\begin{equation}
    \norm {\phi(\mathbf{x}_{1},A)-\phi(\mathbf{x}_{2},A)} \leqslant \norm{\mathbf{x}_{1}-\mathbf{x}_{2}}
\end{equation}
for all $\mathbf{x}_{1}, \mathbf{x}_{2} \in \mathcal{X}$ and $A \in U$; and
\begin{equation}
    \norm{\phi(\mathbf{x},A_{1})-\phi(\mathbf{x},A_{2})} \leqslant L_{2}(1+\norm{\mathbf{x}})\abs{A_{1}-A_{2}}
\end{equation}
for all $\mathbf{x} \in \mathcal{X}$ and $A_{1},A_{2} \in U$. \\

The additional conditions for the SE theorem (see Theorem 1 \emph{(i-ii-iii)} from \cite{rangan2019vector}) to hold are the following:
\begin{itemize}
    \item $\alpha_{1k}$ must be in $[0,1]$. This is always verified using \ref{appendix : prox_prop}, knowing that $f''\geqslant0$ by convexity.
    \item The functions defining $A_{i}$ and $\mathcal{E}_{i}$ must be continuous at the points prescribed by the SE equations. This holds true as well since proximals of convex functions are continuous.
    \item Finally the denoisers (here the proximals) and their derivatives need to be uniformly Lipschitz in their arguments at their parameters. This is again verified from properties of proximal operators and \ref{appendix : prox_prop}.
\end{itemize}
\end{document}